%% file: Neural_ADMM.tex
\documentclass[10pt,twocolumn,letterpaper]{article}

\usepackage{cvpr}
\usepackage{times}
\usepackage{epsfig}
\usepackage{graphicx}
\usepackage{amsmath}
\usepackage{amssymb}
\usepackage{subfigure}
\usepackage{url}
\usepackage{enumerate}   
\usepackage[toc,page]{appendix}
\usepackage{algorithm, algorithmic}

\usepackage[breaklinks=true,bookmarks=false]{hyperref}

\cvprfinalcopy 

\def\cvprPaperID{4046} 

\begin{document}

\graphicspath{{figures/}}

\input notations.tex

\title{InverseNet: Solving Inverse Problems with Splitting Networks}

\author{
Kai Fan\textsuperscript{1}\thanks{The authors contributed equally to this work.}, Qi Wei\textsuperscript{2}$^{\ast}$, Wenlin Wang\textsuperscript{1}, Amit Chakraborty\textsuperscript{2}, Katherine Heller\textsuperscript{1} \\
\textsuperscript{1} Duke University, Durham, North Carolina, USA\\
{\tt\small \{kai.fan, kheller\}@stat.duke.edu, wenlin.wang@duke.edu} \\
\textsuperscript{2} Siemens Corporate Technology, Princeton, New Jersey, USA\\
{\tt\small \{qi.wei, amit.chakraborty\}@siemens.com}
}

\maketitle

\begin{abstract}
We propose a new method that uses deep learning techniques to solve the inverse problems. 
The inverse problem is cast in the form of learning an end-to-end mapping from observed data to the ground-truth. 
Inspired by the splitting strategy widely used in regularized iterative algorithm to tackle inverse problems,
the mapping is decomposed into two networks, with one handling the inversion of the physical forward model 
associated with the data term and one handling the denoising of the output from the former network, i.e., 
the inverted version, associated with the prior/regularization term. The two networks are trained jointly 
to learn the end-to-end mapping, getting rid of a two-step training. The training is annealing as the 
intermediate variable between these two networks bridges the gap between the input (degraded version of output)
and output and progressively approaches to the ground-truth. The proposed network, referred to as \textbf{InverseNet},
is flexible in the sense that most of the existing end-to-end network structure can be leveraged in 
the first network and most of the existing denoising network structure can be used in the second one. 
Extensive experiments on both synthetic data and real datasets on the tasks, motion deblurring, super-resolution, and colorization, demonstrate the efficiency and accuracy of the proposed method compared with other image processing algorithms. 
\end{abstract}

\section{Introduction}
Over the past decades, inverse problems have been widely studied in image and signal processing and computer vision,
e.g., denoising \cite{Elad2006}, deconvolution \cite{Afonso2010}, super-resolution \cite{Zhao2016} and compressive sensing \cite{Figueiredo2007gradient}. An inverse problem is resulted from the forward model which maps unknown signals, 
i.e., the ground-truth, to acquired/observed information about them, which we call data or measurements. 
This forward problem, generally relies on a developed physical theory which reveals the link between the ground-truth and
the measurements. Solving inverse problems involves learning the inverse mapping from the measurements to the ground-truth. 
Specifically, it recovers a signal from one or a small number of degraded or noisy measurements, which
is usually ill-posed \cite{Tikhonov1977}. 
Mathematically, the goal is to reconstruct a high dimensional ground-truth $\mathbf{x}\in \mathbb{R}^{n}$ 
from a low dimensional measurement denoted as $\mathbf{y}\in\mathbb{R}^{m}$, which is reduced from $\mathbf{x}$ by 
a a forward model $A$ such that $\mathbf{y} = A\mathbf{x}$. 
This forward model $A$ is constructed to tie the observed data $\bfy$ to a set of learned model parameters $\bfx$. 
For example, in compressive sensing, $\mathbf{y}$ is a compressive measurement with random sampled regions 
and $A$ is the measurement matrix, e.g., a random Gaussian matrix; in super-resolution, $\mathbf{y}$ is a 
low-resolution image and the operation $A$ downsamples high resolution images. 
The main difficulty of these underdetermined systems comes from the operator $A$ which 
has a non-trivial null space leading to an infinite number of feasible solutions. 
Though most of the inverse problems are formulated directly to the setting of an optimization 
problem associated with the forward model \cite{Tarantola2005inverse}, a number of learning-based algorithms
have been proposed to solve inverse problems by learning a mapping from the measurement domain of $\mathbf{y}$ 
to the signal space of $\mathbf{x}$, with the help of large datasets and neural nets \cite{Yao2004SR,Dong2016PAMI}. 
\footnote{These algorithms refer to directly learning optimum mappings from the observed data to their high-resolution correspondents and are different from learning from training datasets some specific priors to be incorporated in the regularized iterative algorithms\cite{Elad2006,Yang2010}.}

More recently, deep learning techniques have arisen as a promising framework and gained great popularity for providing state-of-the-art performance on applications include pattern analysis (unsupervised), classification (supervised), computer vision, image processing, etc \cite{Deng2014deep}. Exploiting deep neural networks to solve inverse problems has been explored recently \cite{Dong2016PAMI,sonderby2016amortised,adler2017solving,Jin2017}. 
In these works, inverse problems are viewed as a pattern mapping problem and most existing learning-based methods propose to learn an end-to-end mapping from $\mathbf{y}$ to $\mathbf{x}$ \cite{shi2016real,sonderby2016amortised}. 
By leveraging the powerful approximation ability of deep neural networks, these deep learning based data-driven methods have achieved state-of-the-art performance in many challenging inverse problems like super-resolution \cite{bruna2015super,Dong2016PAMI,sonderby2016amortised}, image reconstruction \cite{schlemper2017deep}, automatic colorization \cite{larsson2016learning}. 
More specifically, massive datasets currently enables learning end-to-end mappings from the measurement domain to the target image/signal/data domain to help deal with these challenging problems instead of solving the inverse problem by inference. 
A strong motivation to use neural networks stems from the universal approximation theorem \cite{Csaji2001}, which states that a feed-forward  network with a single hidden layer containing a finite number of neurons can approximate any continuous function on compact subsets of $\mathbb{R}^{n}$, under mild assumptions on the activation function. 
In these recent works \cite{bruna2015super, sonderby2016amortised, larsson2016learning, schlemper2017deep}, an end-to-end mapping from measurements $\bfy$ to ground-truth $\bfx$ was learned from the training data and then applied to the testing data. 
Thus, the complicated inference scheme needed in the conventional inverse problem solver was replaced by feeding a new measurement through the pre-trained network, which is much more efficient. 
However, despite their superior performance, these specifically-trained solvers are designed for specific inverse problems and usually cannot be reused to solve other problems without retraining the mapping function - even when the problems are similar. 
To improve the scope and generability of deep neural network models, more recently, in \cite{chang2017one}, a splitting strategy was proposed to decompose an inverse problem into two optimization problems, where one sub-problem, related to regularization, can be solved efficiently using trained deep neural networks, leading to an alternating direction method of multipliers (ADMM) framework \cite{Boyd2011,7879849}. This method involved training a deep convolutional auto-encoder network for low-level image modeling, which explicitly imposed regularization that spanned the subspace that the ground-truth images lived in. For the sub-problem that required inverting a big matrix, a conventional gradient descent algorithm was used, leading to an alternating update, iterating between feed-forward propagation through a network and iterative gradient descent. 
Thus, an inner loop for gradient descent is still necessary in this framework. 
A similar approach to learn approximate ISTA (Iterative Shrinkage-Thresholding Algorithm) with neural networks was illustrated in \cite{gregor2010learning}. A more flexible splitting strategy of training two reusable neural networks 
for the two sub-problems within ADMM framework leading to an inner-loop free update rule has been proposed 
in \cite{ADMM2017NIPS}. The two pre-trained deep neural networks are flexible and reusable in the sense that the trained network for the proximity operator can be used as a plug-and-play prior to regularize other inverse problems sharing similar statistical characteristics and the trained network for the inversion operator can be used to solve the same inverse problem for other datasets. 

\begin{figure}[t]
\centering 
\subfigure[]{
\includegraphics[width=0.045\textwidth]{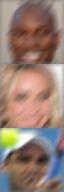}
}
\subfigure[Iter = 1, 25, 50, 100]{
\includegraphics[width=0.09\textwidth]{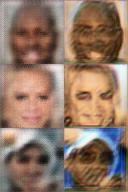}
\includegraphics[width=0.09\textwidth]{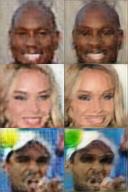}
\includegraphics[width=0.09\textwidth]{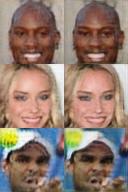}
\includegraphics[width=0.09\textwidth]{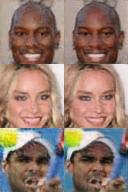}
}
\subfigure[Iter = 200, 400, 800, 1600]{
\includegraphics[width=0.09\textwidth]{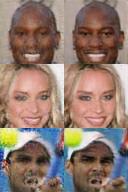}
\includegraphics[width=0.09\textwidth]{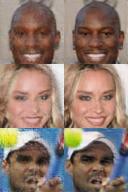}
\includegraphics[width=0.09\textwidth]{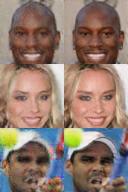}
\includegraphics[width=0.09\textwidth]{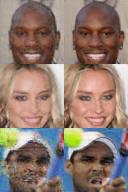}
}
\subfigure[]{
\includegraphics[width=0.045\textwidth]{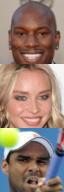}
}
\caption{Illustration of Annealing Training. (a) the motion blurred photos. (b-c) the left column is $\mathbf{z}$ the results of U-Nets and the right column is $\hat{\bfx}$ the results of DAEs. (d) the ground-truth.}
\label{fig:intro}
\end{figure}

To leverage the advantages of both the end-to-end mapping and the splitting strategy, in this work, we propose to
learn an end-to-end mapping consisting of two networks, with one handling the inversion associated with the forward physical model 
and the other one handling the denoising, respectively. A degraded signal, i.e., an observed data point is fed into the inverse network 
to output an intermediate update and then fed into the denoising network to refine the result. The intermediate update bridges the
information gap between the input, i.e., the degraded signal and the output, i.e., the ground-truth. The training for the proposed InverseNet is annealing in the sense that the input of the denoising network, e.g., the denoising autoencoder, referred to as DAE (as an example explained later) or equivalently, the output of the inversion networks, e.g., the U-Nets (as an example explained later), progressively becomes better (closer to the ground-truth) following by a refined (better) result output by the denoising network, as displayed in Fig. \ref{fig:intro}. This training leverages both the data term by the U-Nets and a generative prior by the DAEs. 
More specifically (and heuristically), the inversion network tries to restore the information lost in the forward model, and the denoising network tries to refine the result with learned details, inspired by the two update steps of ADMM.  

{\bf{Contributions:}}
We propose to learn an end-to-end mapping from the observed data to the ground-truth. 
Inspired by the splitting strategy widely used to solve inverse problems as discussed above, instead of learning one network to solve them all, we propose to decompose the end-to-end mapping into two parts with one handling the model inversion part and the other one handling the denoising part. There are several benefits to such structure:
i) Any existing end-to-end learning algorithm can be regarded as the inversion part and incorporated with any existing denoising network. 
ii) The two network can be trained jointly to learn the mapping to its best extent, getting rid of any two-step sub-optimal solution.
iii) In the testing phase, only one feed-forward propagation is necessary to solve the inverse problem, getting rid of any further iterations.
\section{Inverse Problems}
As explained above, the forward model connects the low dimensional measurement $\mathbf{y}\in\mathbb{R}^{m}$ to high dimensional ground-truth $\mathbf{x}\in \mathbb{R}^{n}$ by a linear operator $A$ as $\mathbf{y} = A\mathbf{x}$. The fact that $n \geq m$ makes the number of parameters to estimate larger than the number of available data points in hand. 
Since $A$ is an underdetermined measurement matrix, this imposes an ill-posed problem for finding solution $\mathbf{x}$ on a new 
observation $\mathbf{y}$. For instance, the matrix $A$ is a strided Gaussian convolution and not invertible for super-resolution tasks in \cite{shi2016real, sonderby2016amortised}. To address this issue, computational solutions including approximate inference based on Markov chain Monte Carlo (MCMC) and optimization based on variable splitting under the ADMM framework, were proposed and applied to different kinds of priors, e.g., the empirical Gaussian prior \cite{Wei2015JSTSP,Zhao2016}, the Total Variation prior \cite{Simoes2015}, etc. The ADMM framework is popular due to its low computational 
complexity and recent success in solving large scale optimization problems. 
Mathematically, the optimization problem is formulated as 
%
\begin{align}\label{eq:ADMM}
\hat{\mathbf{x}} = \arg\min_{\mathbf{x}, \mathbf{z}} \|\mathbf{y} - A\mathbf{z}\|^2 + \lambda\mathcal{R}(\mathbf{x}), \quad s.t. \quad \mathbf{z} = \mathbf{x}
\end{align}
where the introduced auxiliary variable $\bfz$ is forced to be equal to $\bfx$, and $\mathcal{R}(\mathbf{x})$ models the structure promoted by the prior/regularization. To leverage the befinits of `big data', the regularization can be imposed in an empirical Bayesian way, by designing an implicit data dependent prior on $\mathbf{x}$, i.e., $\mathcal{R}(\mathbf{x}; \mathbf{y})$ for amortized inference \cite{sonderby2016amortised}. The augmented Lagrangian for Eq.~(\ref{eq:ADMM}) by replacing $\mathcal{R}(\mathbf{x})$ with 
$\mathcal{R}(\mathbf{x}; \mathbf{y})$ is
\begin{equation}
\begin{split}
&\mathcal{L}(\mathbf{x}, \mathbf{z}, \mathbf{u}) = \\
 &\|\mathbf{y} - A\mathbf{z}\|^2 + \lambda\mathcal{R}(\mathbf{x}; \mathbf{y}) + \langle \mathbf{u}, \mathbf{x} - \mathbf{z} \rangle + \beta \|\mathbf{x} - \mathbf{z}\|^2
\end{split}
\end{equation}
where $\mathbf{u}$ is the Lagrange multiplier and $\beta > 0$ is the penalty parameter. 
The conventional augmented Lagrange multiplier method that minimizes $\mathcal{L}$ w.r.t. $\mathbf{x}$ and $\mathbf{z}$ simultaneously,
is difficult and does not exploit the fact that the objective function is separable. To tackle this problem, ADMM decomposes the minimization into two subproblems, i.e., minimizations w.r.t. $\mathbf{x}$ and $\mathbf{z}$, respectively. 
More specifically, the updates are as follows: 
\begin{align}
\mathbf{z}^{k+1} &= \arg\min_{\mathbf{z}} \|\mathbf{y} - A\mathbf{z}\|^2 + \beta \| \mathbf{x}^{k+1} - \mathbf{z} + \frac{\mathbf{u}^k }{2\beta} \|^2 \label{eq:minz}\\
\mathbf{x}^{k+1} &= \arg\min_{\mathbf{x}} \beta \| \mathbf{x} - \mathbf{z}^k + \mathbf{u}^k/2\beta\|^2 + \lambda\mathcal{R}(\mathbf{x}; \mathbf{y}) \label{eq:minx}\\
\mathbf{u}^{k+1} &= \mathbf{u}^k + 2\beta(\mathbf{x}^{k+1} - \mathbf{z}^{k+1}). 
\label{eq:updateu}
\end{align}
For the priors $\mathcal{R}$ of special forms, such as $\|\mathbf{x}\|_1$, a closed-form solution for Eq.~$\eqref{eq:minx}$, i.e., a soft thresholding solution is easily obtained. On the contrary, for some more sophiscated regularizations, e.g., a patch based prior \cite{Elad2006,Yang2010}, solving Eq.~\eqref{eq:minx} is nontrivial, and may require iterative methods. To solve Eq.~\eqref{eq:minz}, a matrix inversion is inevitable, for which gradient descent (GD) method is usually applied to update $\mathbf{z}$ \cite{chang2017one}. 
Thus, solving Eq.~\eqref{eq:minx} and \eqref{eq:minz} is in general cumbersome as inner loops are required to solve these two sub-minimization problems. 
More specifically, most of computational complexity comes from the proximity operator due to its intractability and the matrix inversion which is not easy diagonalized. \\
{\bf{Motivation to introduce ADMM}}:
The proposed network structure is inspired by the two iterative steps in ADMM updates to solve an inverse problem. 
More specially, the inverse network imitates the process of solving Eq.~\eqref{eq:minz} and the denoising network imitates 
an algorithm to solve Eq.~\eqref{eq:minx}. 
In this work, the inverse network exploits the U-Nets structure \cite{ronneberger2015u} and the denoising network 
uses the DAE \cite{Vincent2008}, which will be elaborated in the following section.

\begin{figure*}
\centering 
\includegraphics[width=\textwidth]{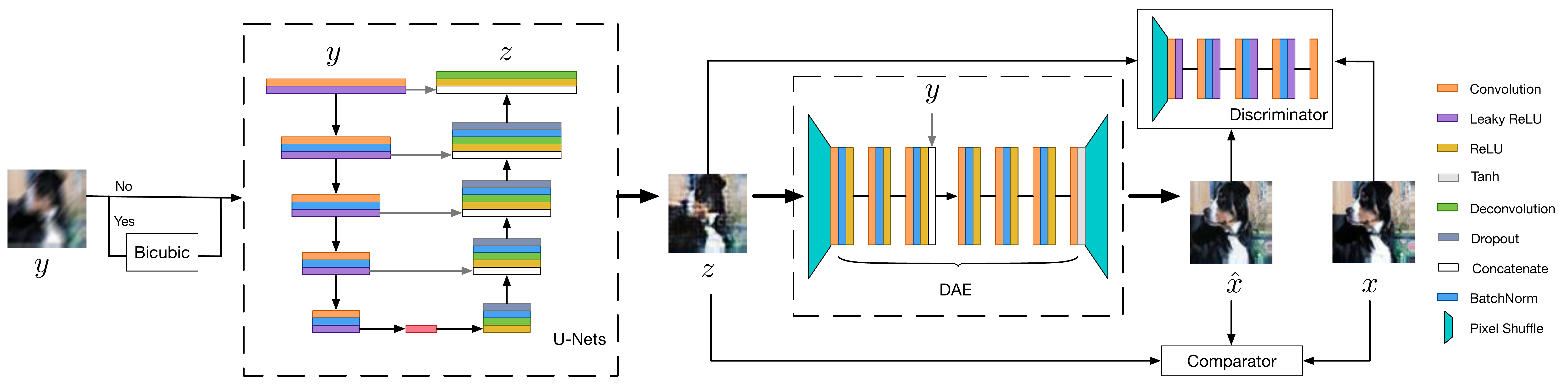}
\caption{The pipeline architecture of \emph{InverseNet}. Initially, the degraded input $\mathbf{y}$ is pre-processed by bicubic resizing depending on the task, e.g., for super-resolution. Thus, the input $\mathbf{y}$ of U-Nets has the same size as the output $\mathbf{z}$. Next, the output $\mathbf{z}$ is fed into the DAEs with two pixel shuffling blocks. In the end, the intermediate result $\mathbf{z}$ and final estimation $\hat{\mathbf{x}}$ are both fed into the discriminator and \emph{comparator} with the ground-truth.}
\label{fig:model}
\end{figure*}

\section{Approach}
The proposed method decomposes the end-to-end mapping from data to ground-truth into two mappings,
one corresponding to inversion of the physical model and the other corresponding to the regularized denoising. 
From a Bayesian perspective, the first mapping handles the likelihood associated with data term 
and the second mapping handles the prior term. 
As shown in Fig. \ref{fig:model}, the observed data $\bfy$ is fed into the first network to 
get an approximation $\mathbf{z}$ of the ground-truth. 
Note that $\mathbf{z}$ is of the same size as $\bfx$ and can be regarded as a noisy version of it. 
The noisy one $\mathbf{z}$ is then fed into the second network to get refined to the clear version $\hat{\bfx}$. 
This two-network structure is also echoed from the recent popular refining technique used in \cite{shankar2016refining,chen2017photographic}.
More details about these two networks will be elaborated in the following sections.

\subsection{Inversion network mapping $\bfy$ to $\mathbf{z}$}
The inversion network tries to learn the inversion of the degradation which maps the ground-truth $\bfx$ to the observed data $\bfy$. 
Note that while it may be straightforward to write down the closed-form solution for sub-problem \ref{eq:minz} w.r.t. $\bfz$, explicitly computing this solution is nontrivial due to the indefeasibly of inverting a big matrix. 
Similar to the strategy in \cite{ADMM2017NIPS,sonderby2016amortised}, we design a deep convolutional neural network to learn the inversion. 
More specifically, we have used a recently developed network structure referred to as U-Nets \cite{ronneberger2015u}, which was originally developed for medical image segmentation. 
The U-Net architecture allows low-level information to shortcut by filtering concatenation across different layers, differing from the element-wise addition in ResNet \cite{He_2016_CVPR}. 
Extensive experiments have demonstrated the effectiveness of U-Nets to learn the complex mapping between high-dimensional data, such as images \cite{cciccek20163d,Jin2017}. 

In general, U-Nets require the input and output to have the same size, for the height and width at least. 
However, in many inverse problems, for example, super-resolution or compressive sensing, only a low dimensional signal 
is available as the input. 
In such cases, we initially apply the bicubic interpolation to obtain the same size input which is suitable for the U-Nets. 
As a summary, the architecture of our U-Nets is shown in the left dashed box in Fig. \ref{fig:model}. 

\subsection{Denoising network mapping $\mathbf{z}$ to $\hat{\bfx}$}
The denoising network plays a role to learn from the dataset a signal prior that can deal with any inverse problems. 
In optimization algorithms for solving inverse problems, signal priors are usually cast in the form of a denoiser, more specifically, a proximity operator \cite{Combettes2011}. 
From the geometrical perspective, the proximity operator projects a noisy data point into the feasible sets spanned by the signal prior. 
Thus, any existing denoising network can be exploited in the proposed framework. 
In our work, we propose to a specially designed denoising auto-encoders with pixel shuffling (or sub-pixel) trick \cite{shi2016real}. 
Similar to the inversion net, pixel shuffling convolutional network is designed to deal with inputs and outputs of the same size by periodically reordering the pixels in each channel mapping a high resolution image to to the scale as the same as the low dimensional image. 
The resulting denoising auto-encoders does not have a bottle-neck shape structure as each layer shares the same filter size. 
This allows us to transform $\mathbf{z}$ into a tensor that has the same size as $\bfy$, and concatenate this tensor with $\bfy$ 
if desired in the regularization of amortized inference. The detailed architecture of the proposed DAE shown in Fig. \ref{fig:model}. 

\subsection{Practical Adversarial Training}

From either the perspective of the model structure or the ADMM update rules, $\mathbf{z}$ and $\hat{\mathbf{x}}$ are both 
approximates of the ground-truth $\mathbf{x}$. Inspired by the idea in autoencoding beyond pixels using a learned similarity metric \cite{larsen2016autoencoding}, standard negative likelihood loss (equivalent to reconstruction loss) incorporating additional generative adversarial loss can practically improve the quality of generated samples. 
In our work, we also force the outputs of two designed networks to share one discriminator used as a binary classifier, inducing the standard minimax loss \cite{goodfellow2014generative} in our model, i.e.,
\begin{align}
\mathcal{L}_D &= \mathcal{L}_{GAN:D}(\mathbf{z}) + \mathcal{L}_{GAN:D}(\hat{\mathbf{x}}) \label{eq:loss_d} \\
\mathcal{L}_G & = \mathcal{L}_{GAN:G}(\cdot) + \lambda_l\mathcal{L}_{likelihood}(\cdot) \label{eq:loss_g1},  
\end{align}
where $\mathcal{L}_{GAN:D}$ is the negative likelihood loss with respect to classification problem, $\mathcal{L}_{GAN:G}$ is the non-saturated generative loss, and ``$\cdot$'' in Eq.~(\ref{eq:loss_g1}) should be substituted by $\mathbf{z}$ or $\hat{\mathbf{x}}$ to introduce two different generative loss functions.

Besides using the classifier as the discriminator, many works \cite{salimans2016improved, arjovsky2017wasserstein} argued that 
a pre-trained regression network to match the high level features between the real images and the fake ones can be significantly 
better than the discriminative loss in practice and theoretically is more robust from the viewpoint of distribution matching. 
Therefore, a regression network referred to as \emph{comparator} is shared by the outputs of two proposed nets as well. 
Since the transfer learning is well established in the image domain, such as style learning with feature matching \cite{gatys2015neural} in VGG \cite{simonyan2014very}, we prefer to use the pre-trained VGG or AlexNet \cite{krizhevsky2012imagenet} as the \emph{comparator}. 
In this case, to leverage both the adversarial training and the \emph{comparator}, we do not modify the discriminator loss in Eq.~(\ref{eq:loss_d}), but add an extra feature matching loss to Eq.~(\ref{eq:loss_g1}) as
\begin{align}
\mathcal{L}_G & = \mathcal{L}_{GAN:G}(\cdot) + \lambda_l\mathcal{L}_{likelihood}(\cdot) + \lambda_f \mathcal{L}_{feature}(\cdot). \label{eq:loss_g2}
\end{align}
Analogous to the traditional optimization for generative adversarial loss, 
$\mathcal{L}_D$ and $\mathcal{L}_G$ are iteratively minimized with respect to the parameters of the discriminator, U-Nets, and DAEs.  

\subsection{Underlying Annealing Training}
The proposed adversarial training leads to an underlying annealing training for DAEs where the input (the output of U-Nets) at early training stage is extremely noisy, and then progressively approaches to a clearer version with further training. 
This incremental tuning nature in adversarial training allows the training to imitate the inner loop of ADMM updates, 
where the difference $\mathbf{u}$ in Eq.~(\ref{eq:updateu}) gradually becomes constant and negligible. 
This is the main reason why we call our model as \textit{InverseNet}. 
Unlike the traditional DAEs, in which the input data is contaminated by the pre-defined noise, e.g., Gaussian noise with fixed variance,
the output of U-Nets or the input of DAEs in the proposed network is contaminated by non-stationary and time-variant noise without an explicit analytic form. If we remove the DAEs part from our model, we found that the performance gradually becomes worse 
during the training, which can be explained by the instability of adversarial training \cite{arjovsky2017wasserstein}. 
The existence of the DAEs make the network be able to purify the complicated noisy output from the inversion network, leading to a much stable and efficient training. 


\textbf{Generality}
A critical point for learning-based methods is whether the method generalizes to other problems. 
More specifically, how does a method that is trained on a specific dataset perform when applied to another dataset? 
To what extent can we reuse the trained network without re-training? Compared with the reusable matrix inversion neural networks learned with pure noise proposed in \cite{sonderby2016amortised, ADMM2017NIPS}, this inversion network is less flexible due to the joint training of the inversion and denoising network. However, because the learning is based on problem formulation, though not fully reusable, the well-trained neural networks have better generalization ability compared with the other end-to-end learning networks.  
For example, the well-trained networks for image motion deblurring on PASCAL VOC dataset \cite{everingham2010pascal} 
can be directly applied to the same task on ImageNet \cite{deng2009imagenet} with no necessity of fine-tuning, if the degrade operator $A$ remains the same, since these two datasets are both natural images. However, if a significant domain changes to the dataset, such as MRI, all networks-reusable algorithms will fail and require re-training as well as our approach. 

\section{Experiments and Results}

In this section, we provide experimental results and analysis on the proposed \emph{InverseNet} for solving inverse problems. 
The datasets used in our experiments include Caltech-UCSD Birds-200-2011 (CUB) \cite{wah2011caltech},
Large-scale CelebFaces Attributes (CelebA) \cite{liu2015deep}, PASCAL VOC and ImageNet \cite{krizhevsky2012imagenet},
and the learning tasks considered include motion deblurring, $\times4$ super-resolution, and joint $\times2$ super-resolution and colorization. Note that the degraded operator $A$ in unknown during training, and is only used to generate the measurement and ground-truth paired data. Our code will be available on the repository \url{https://github.com/}.

\subsection{Motion Deblurring}

\begin{table}
\begin{center}
\setlength{\tabcolsep}{3pt}
\begin{tabular}{|l|c|c|c|}
\hline
PSNR & CUB & CelebA & ImageNet \\
\hline
Wiener filter (baseline) & 22.42 & 20.92 & 20.44 \\
Robust motion deblur \cite{xu2010two} & 25.03 & 25.06 & 23.37 \\
Neural motion deblur \cite{chakrabarti2016neural} & 25.73 & 25.76 & 24.74 \\
Pix2Pix \cite{isola2016image} w. \emph{comparator} & 23.67 & 23.59 & 22.05 \\
\hline
Ours & 28.39 & 34.02 & 28.87 \\
\hline
\hline
SSIM & CUB & CelebA & ImageNet \\
\hline
Wiener filter (baseline) & 0.6572 & 0.7020 & 0.6357 \\
Robust motion deblur \cite{xu2010two} & 0.7459 & 0.8052 & 0.7283  \\
Neural motion deblur \cite{chakrabarti2016neural} & 0.8853 & 0.9649 & 0.9074 \\
Pix2Pix \cite{isola2016image} w. \emph{comparator} & 0.7554 & 0.8553 & 0.7335 \\
\hline
Ours & 0.9421  & 0.9738 & 0.9446  \\
\hline
\end{tabular}
\end{center}
\caption{PSNR and SSIM Comparison on motion deblurring}
\label{tab:motion_deblur}
\end{table}

\begin{figure*}
\centering
\subfigure[Blurred image]{
\includegraphics[width=0.159\textwidth]{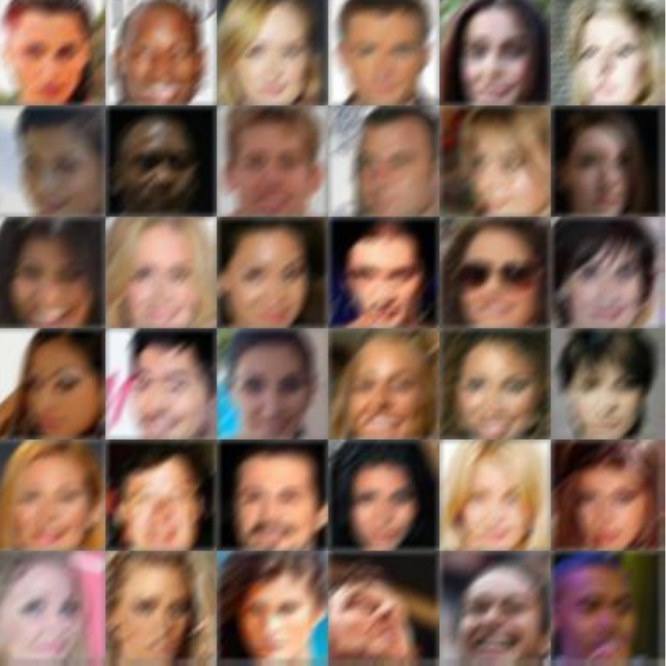}
\includegraphics[width=0.159\textwidth]{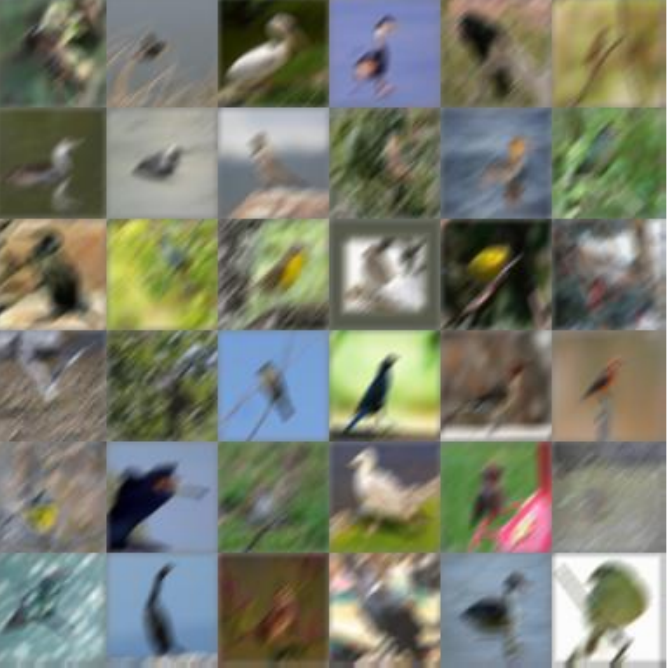}
\includegraphics[width=0.159\textwidth]{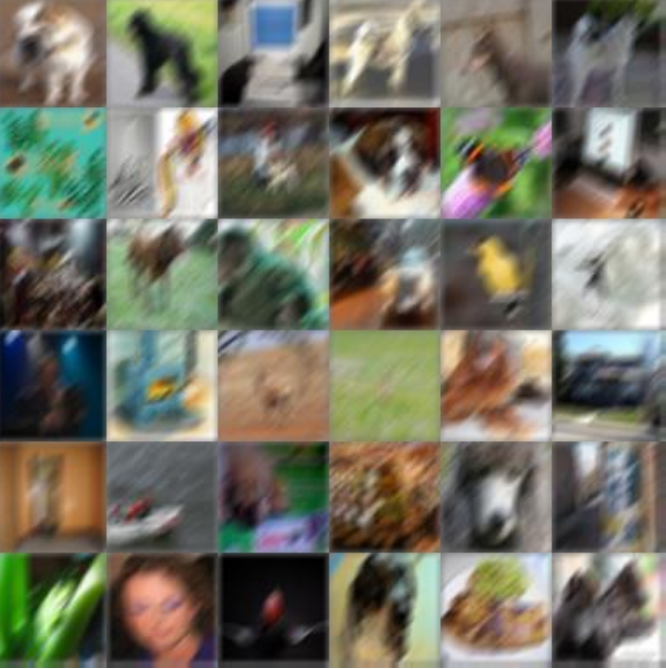}
}
\subfigure[Baseline: Wiener filter]{
\includegraphics[width=0.159\textwidth]{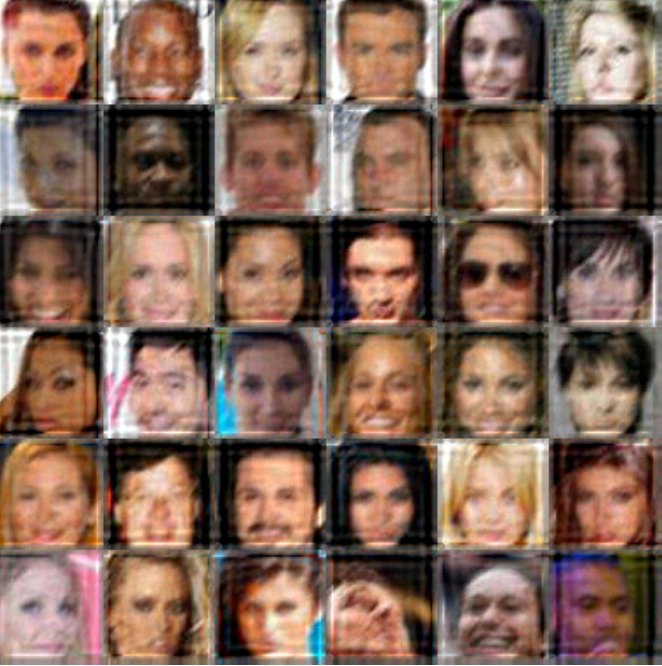}
\includegraphics[width=0.159\textwidth]{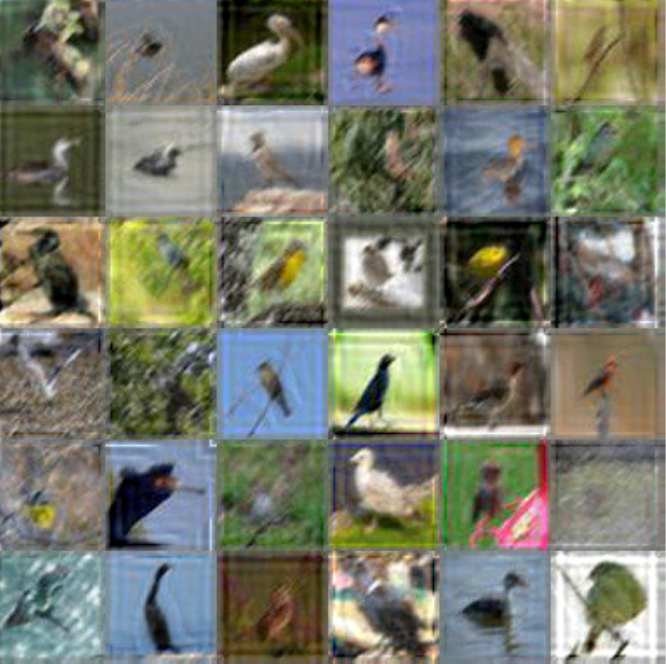}
\includegraphics[width=0.159\textwidth]{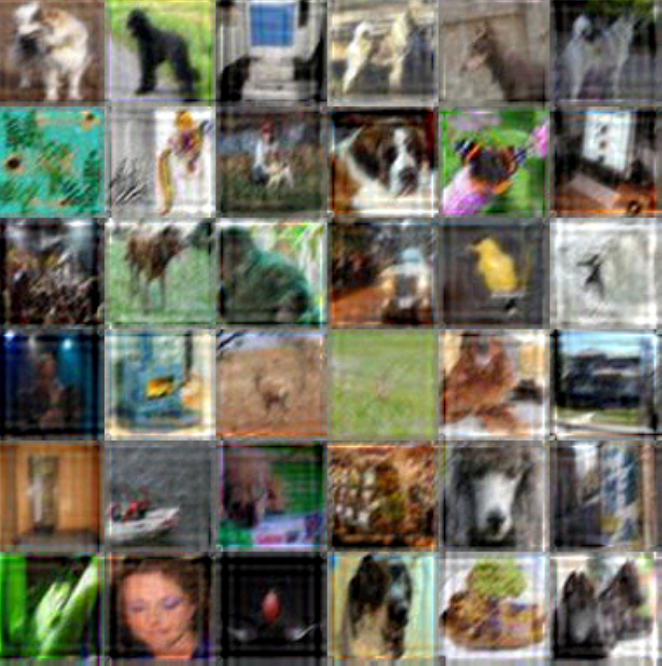}
}
\subfigure[Robust motion deblurring \cite{xu2010two}]{
\includegraphics[width=0.159\textwidth]{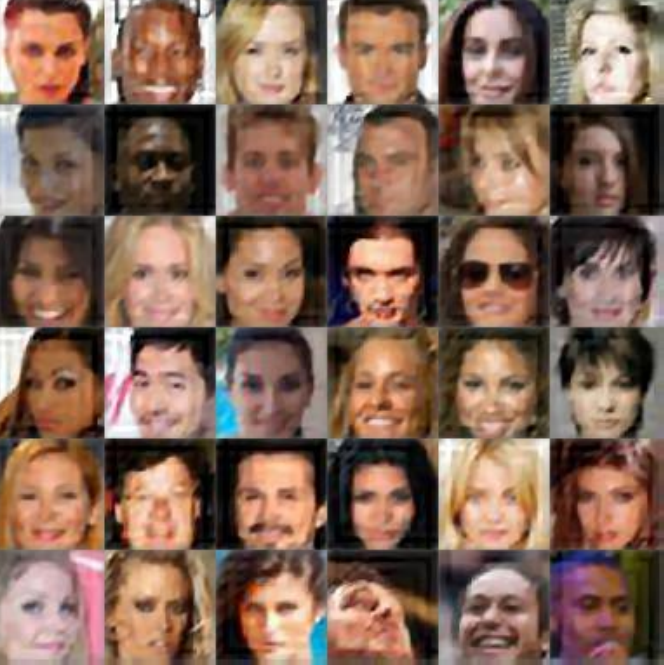}
\includegraphics[width=0.159\textwidth]{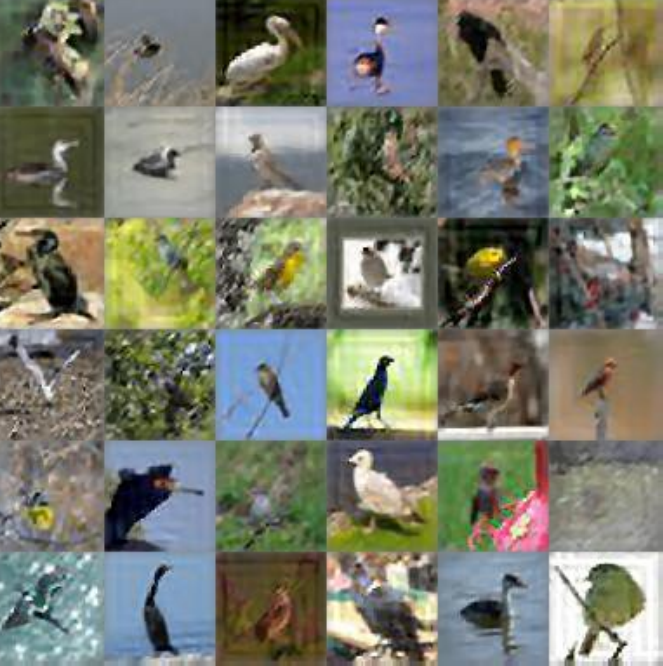}
\includegraphics[width=0.159\textwidth]{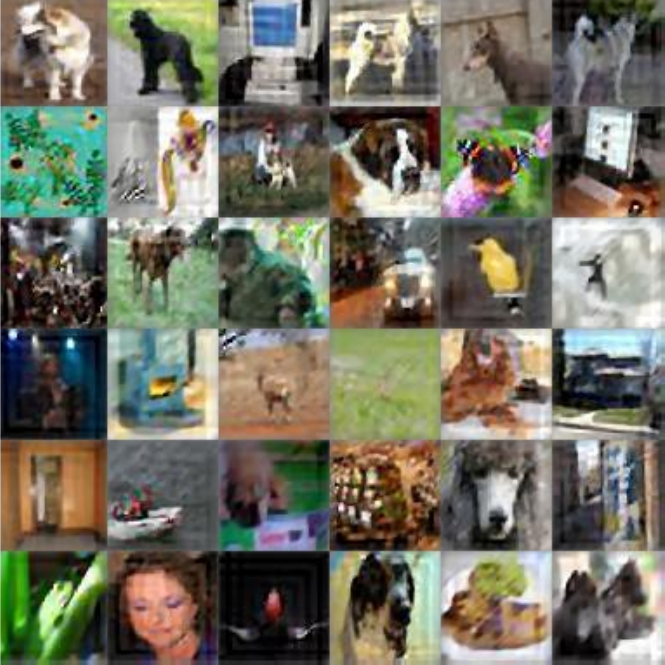}
}
\subfigure[Neural motion deblurring \cite{chakrabarti2016neural}]{
\includegraphics[width=0.159\textwidth]{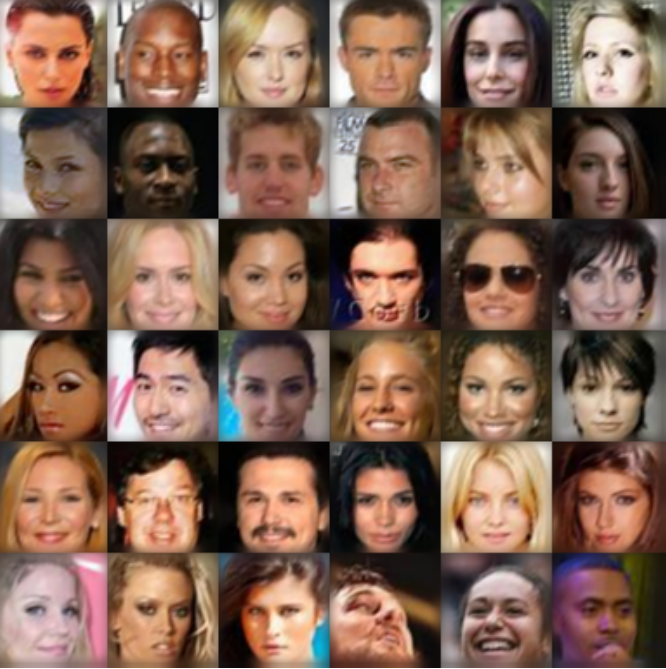}
\includegraphics[width=0.159\textwidth]{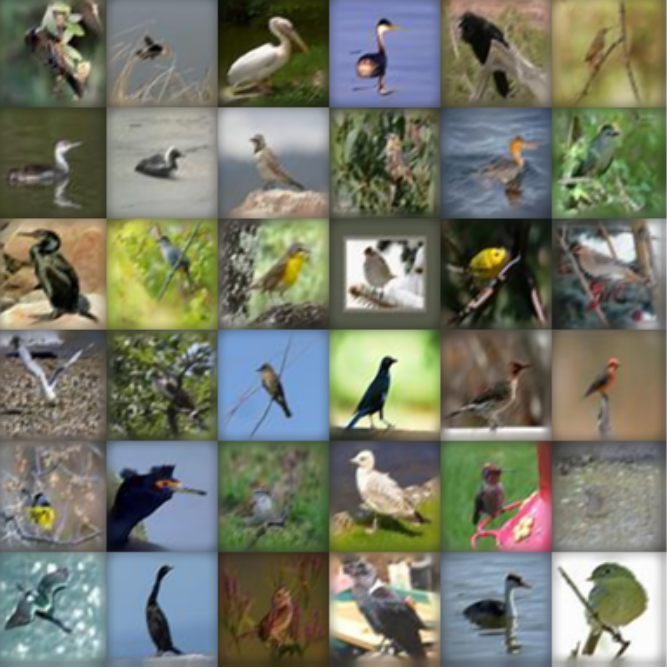}
\includegraphics[width=0.159\textwidth]{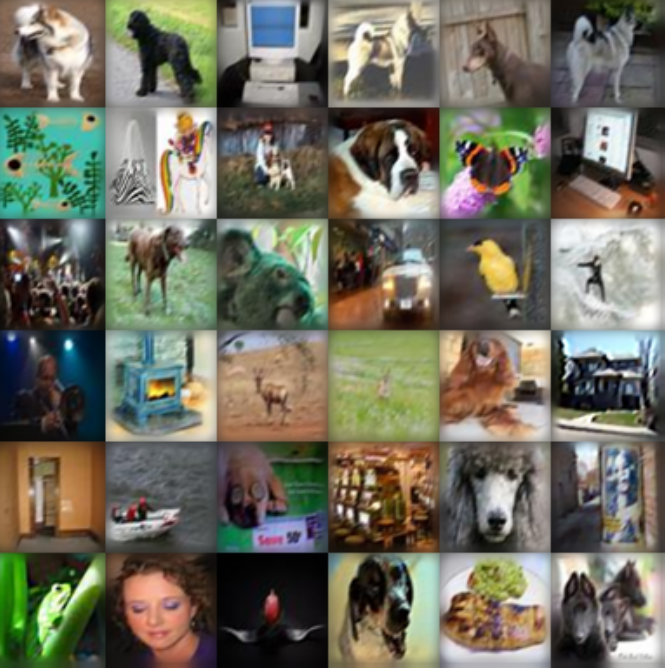}
}
\subfigure[Ours: InverseNet]{
\includegraphics[width=0.159\textwidth]{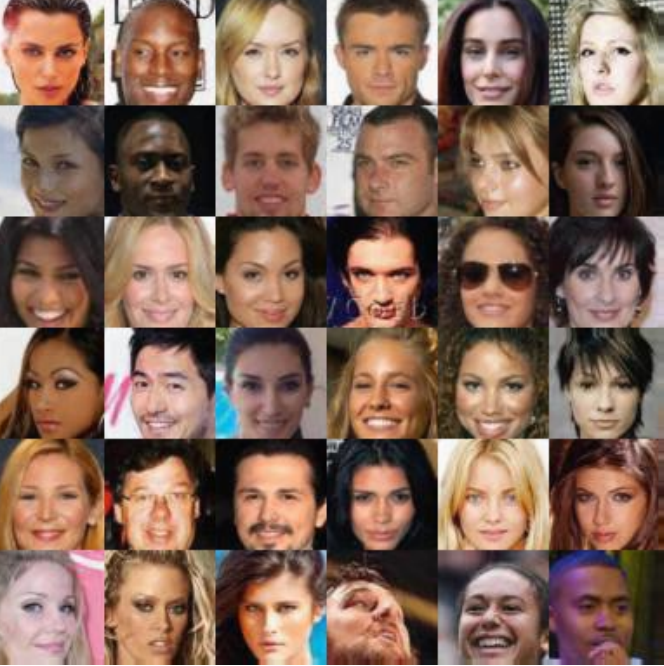}
\includegraphics[width=0.159\textwidth]{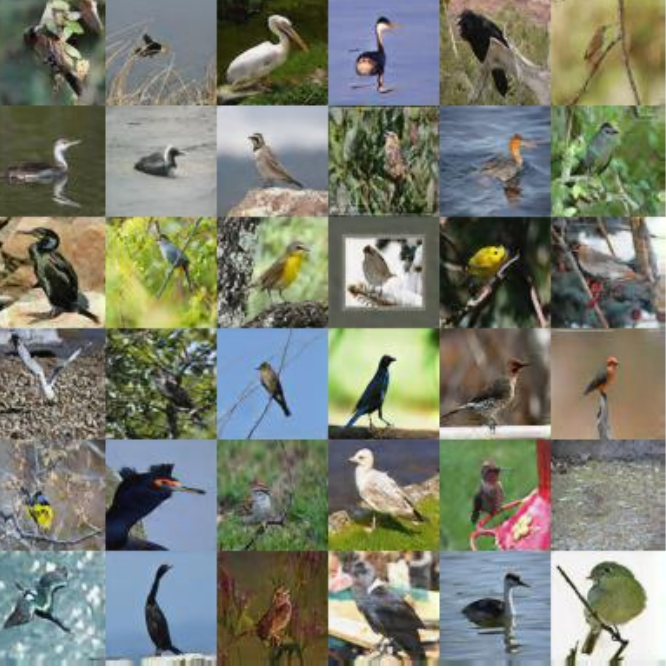}\
\includegraphics[width=0.159\textwidth]{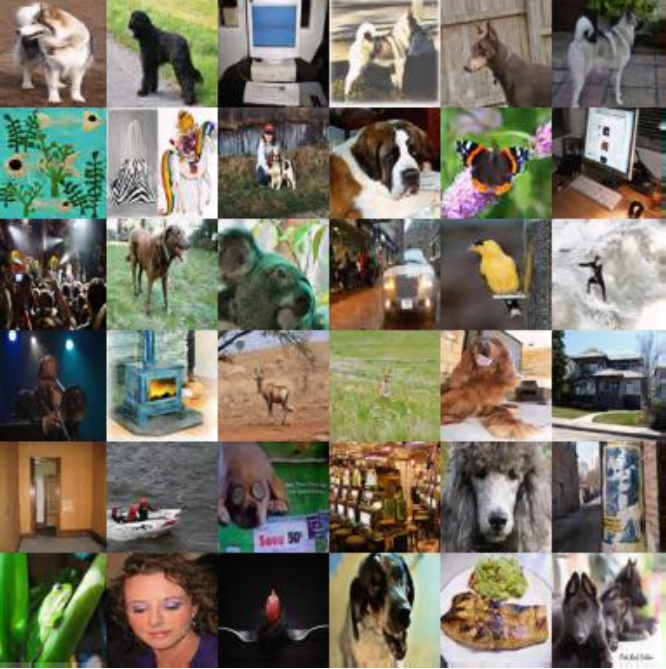}
}
\subfigure[Ground-truth]{
\includegraphics[width=0.159\textwidth]{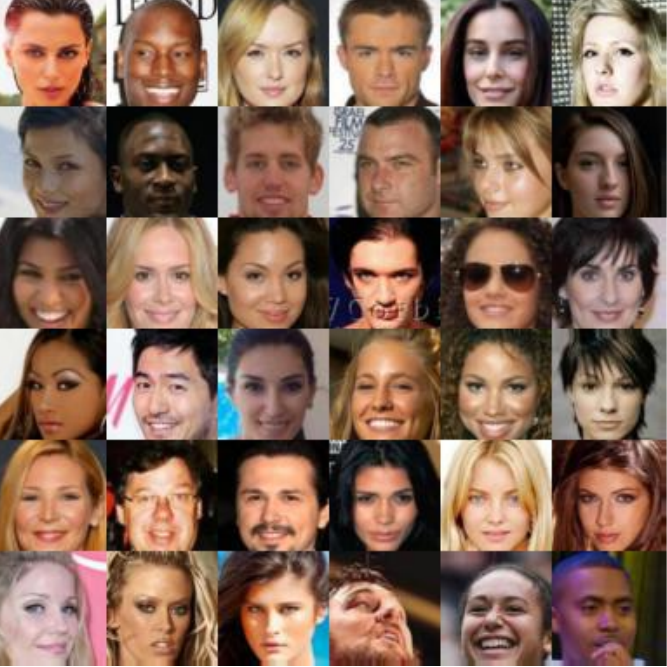}
\includegraphics[width=0.159\textwidth]{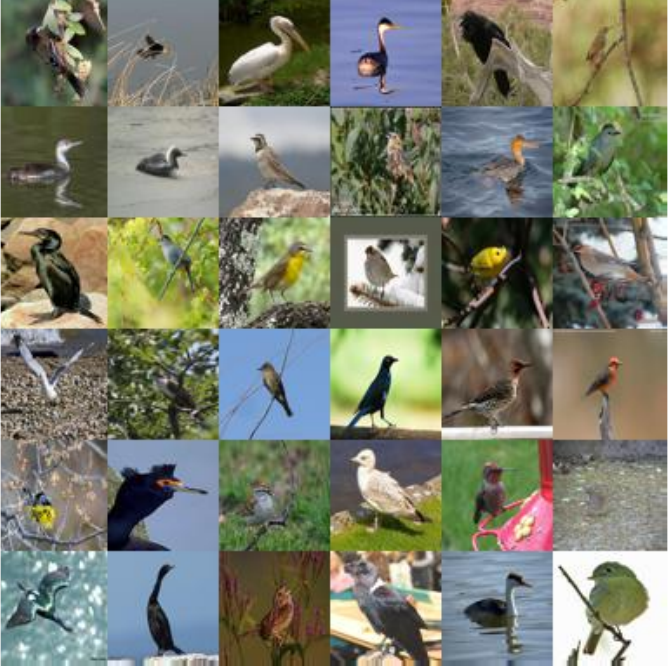}
\includegraphics[width=0.159\textwidth]{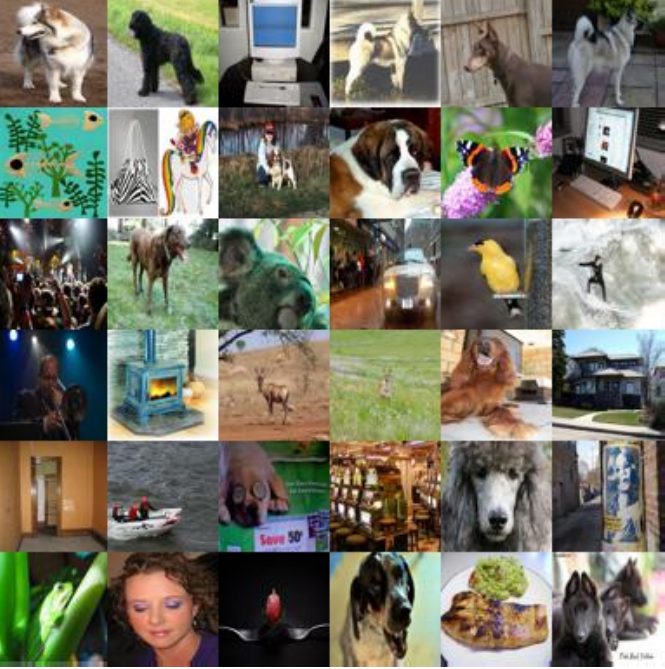}
}
\caption{Example deblurred results from different methods and three different datasets (zooming in can see details).}
\label{fig:motion_deblur}
\end{figure*} 

\begin{figure}
\centering
\includegraphics[width=0.115\textwidth]{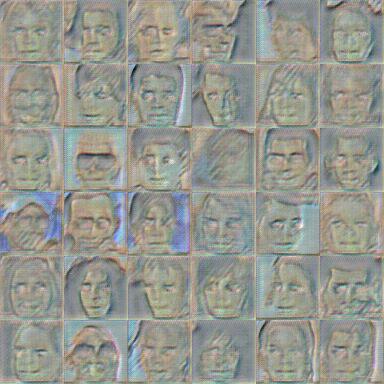}
\includegraphics[width=0.115\textwidth]{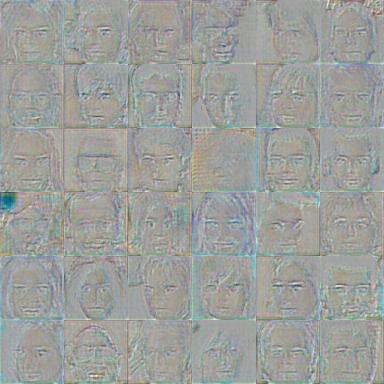}
\includegraphics[width=0.115\textwidth]{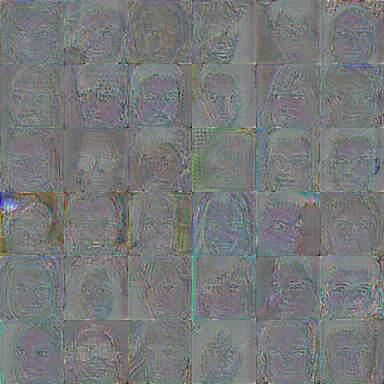}
\includegraphics[width=0.115\textwidth]{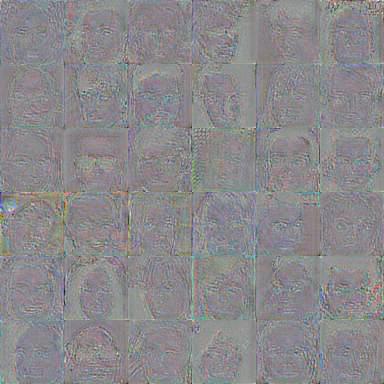}
\includegraphics[width=0.115\textwidth]{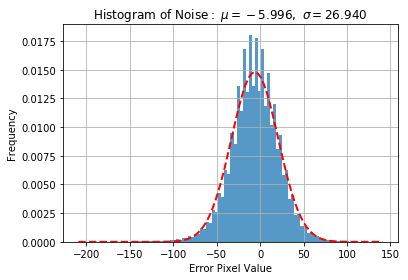}
\includegraphics[width=0.115\textwidth]{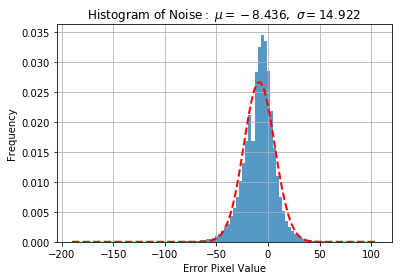}
\includegraphics[width=0.115\textwidth]{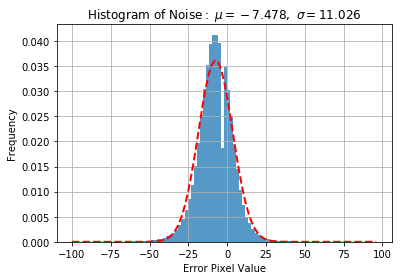}
\includegraphics[width=0.115\textwidth]{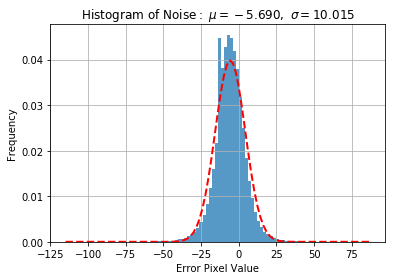}
\includegraphics[width=0.115\textwidth]{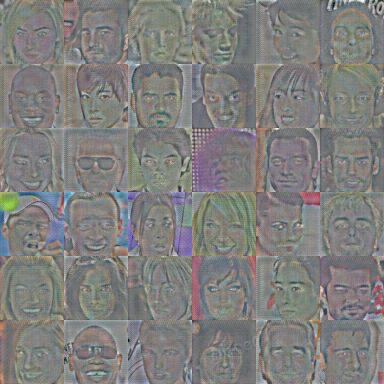}
\includegraphics[width=0.115\textwidth]{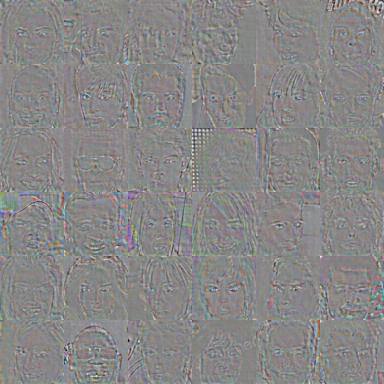}
\includegraphics[width=0.115\textwidth]{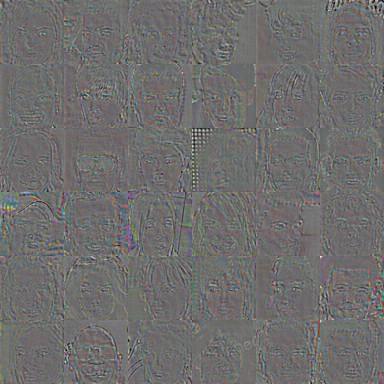}
\includegraphics[width=0.115\textwidth]{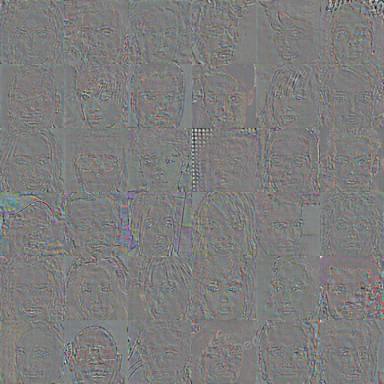}
\includegraphics[width=0.115\textwidth]{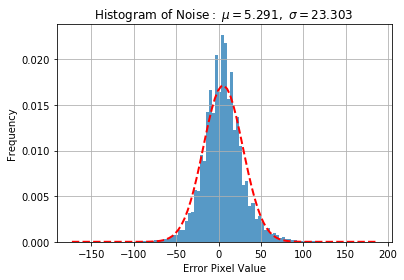}
\includegraphics[width=0.115\textwidth]{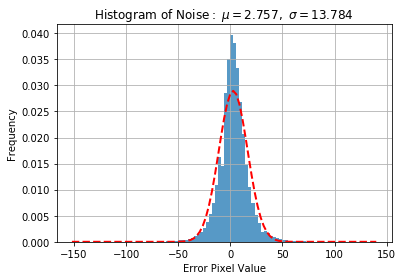}
\includegraphics[width=0.115\textwidth]{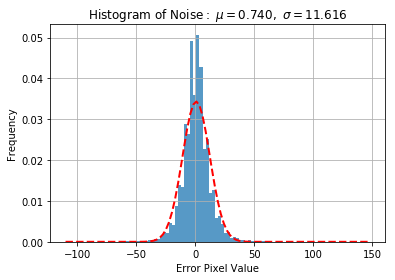}
\includegraphics[width=0.115\textwidth]{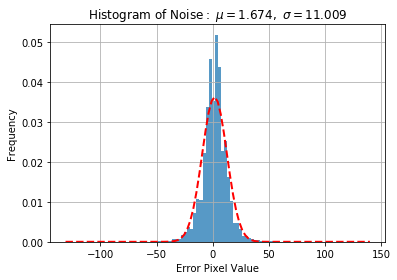}
\caption{Annealing training at iteration 1, 25, 50, 100 on celebA dataset. 1st and 3rd Rows: the error images between $\mathbf{z}$ and model estimation $\hat{\mathbf{x}}$, and the ones between $\mathbf{z}$ and ground-truth $\mathbf{x}$. 2nd and 4th Rows: the annealing empirical distribution of pixel level noise w.r.t $\bar{\mathbf{x}}-\mathbf{z}$ and $\mathbf{x}-\mathbf{z}$.}
\label{fig:anneal}
\end{figure}

\begin{figure}
\centering
\includegraphics[width=0.156\textwidth]{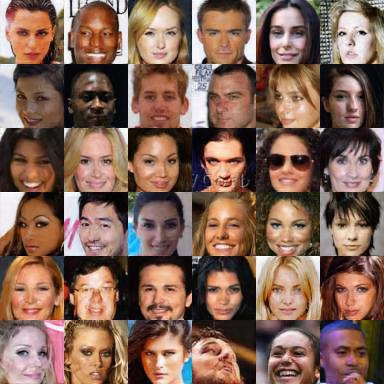}
\includegraphics[width=0.156\textwidth]{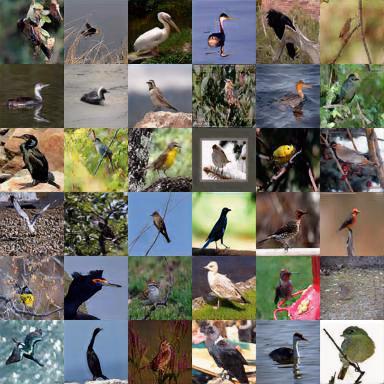}
\includegraphics[width=0.156\textwidth]{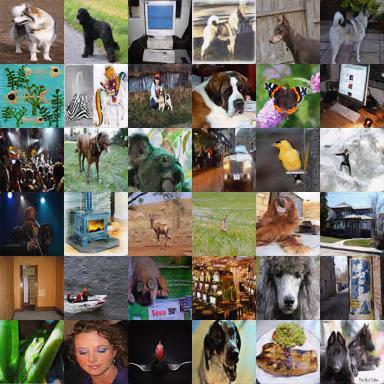}
\caption{\textbf{Transfer Learning}: Testing on the same images from Fig.~\ref{fig:motion_deblur}(a) with well-trained model on VOC dataset.}
\label{fig:adaption}
\end{figure}

The target of motion deblurring is to recover a sharp non-blurred image from a single motion-blurred image, 
which has been a fundamental problem in computational imaging \cite{xu2010two}. The motion deblurring is challenging 
as both the blurring kernel and the latent non-blurred images are unknown in most cases, leading to an ill-posed inverse problem. 
Note that the size of an image after motion deblurring remains the same. Thus, any resizing of the input images, 
e.g., bicubic interpolation or pixel shuffling for the initialization of our approach is unnecessary and 
can be simply removed from the pipeline. We trained our model on all datasets with size $64 \times 64$ images 
without any other pre-processing. 

\textbf{Qualitative Results} The proposed \emph{InverseNet} is compared with the baseline optimization methods 
by Wiener filtering, the robust non-blind motion deblurring \cite{xu2010two}, and another deep learning approach method neural motion deblurring \cite{chakrabarti2016neural} visually in Fig.~\ref{fig:motion_deblur}. Note that in \cite{xu2010two}, 
the algorithm for a single image motion deblurring requires to feed the blurring kernel, while other methods 
including ours are all blind recoveries. As shown in Fig.~\ref{fig:motion_deblur}, the traditional single image 
processing algorithms including Wiener filtering and robust motion deblurring suffered heavily from ring artifacts
while the neural motion deblurring and the proposed \emph{InverseNet} gave much better and high-quality recoveries. 
The ring artifact can be explained by the fact that the chosen convolutional kernel to blur images has larger variance and is more challenging. 

\textbf{Quantitative Results} The PSNR and SSIM were calculated on the same testing datasets across all compared algorithms
and summarized in Table.~\ref{tab:motion_deblur}. Clearly, the proposed \emph{InverseNet} outperformed the other
methods with a significant improvement, which is in consistence with the qualitative results 
in Fig. \ref{fig:motion_deblur}. 

\textbf{Transfer Training}
We also checked to what extent the well-trained model could be adapted to other datasets. Specifically, we trained the \emph{InverseNet} on PASCAL VOC and tested it on CUB, celebA and ImageNet, with the results shown in Fig.~\ref{fig:adaption}. 
The quantitative results is shown in Table~\ref{tab:transfer}. As we argued before, the similar image domain between ImageNet and VOC leads the PSNR does not decrease much, i.e., from 28.87 to 28.52, compared with using the InverseNet trained on ImageNet itself. 
This demonstrated the generality of the learned \emph{InverseNet}.

\begin{table}
\begin{center}
\setlength{\tabcolsep}{3pt}
\begin{tabular}{|l|c|c|c|}
\hline
Metric & CUB & CelebA & ImageNet \\
\hline
PSNR & 25.65 & 27.90 & 28.52 \\
SSIM & 0.9105 & 0.9373 & 0.9303 \\
\hline
\end{tabular}
\end{center}
\caption{Performance with well-trained model on VOC}
\label{tab:transfer}
\end{table}

\begin{figure*}[ht]
\centering
\includegraphics[width=\textwidth]{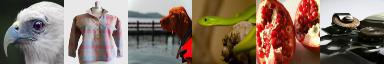}
\includegraphics[width=\textwidth]{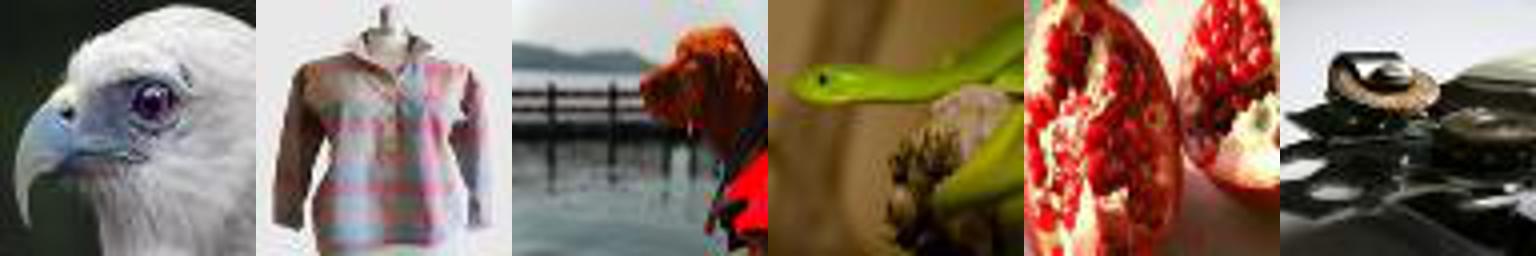}
\includegraphics[width=\textwidth]{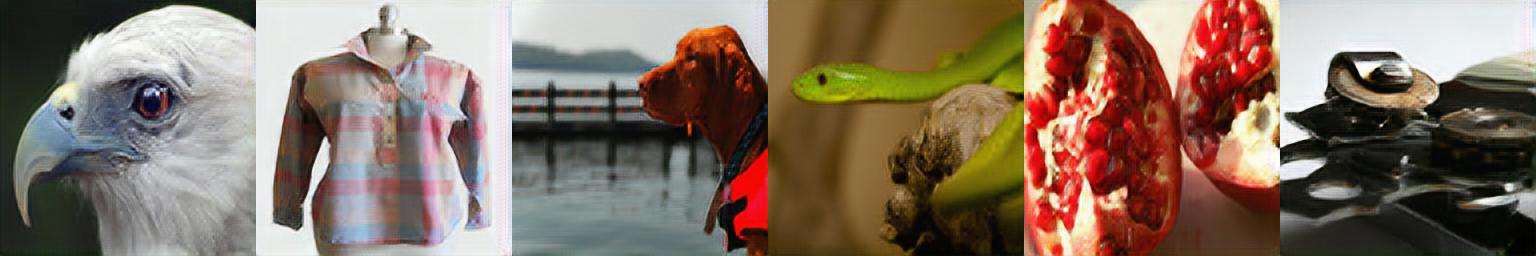}
\includegraphics[width=\textwidth]{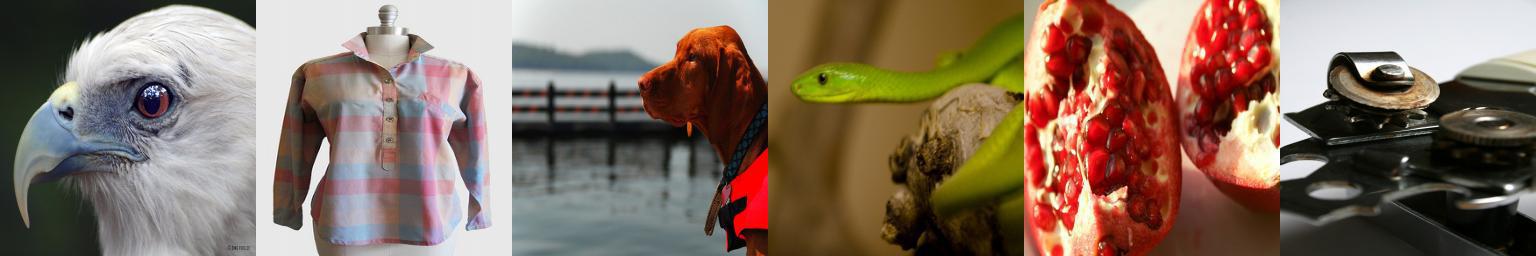}
\caption{\textbf{Ring Effect Remover} on ImageNet for $64\times 64\rightarrow 256\times 256$. 1st Row: LR images; 2nd Row: The bicubic interpolation results (having ring artifacts); 3rd Row: Results by \emph{InverseNet}; 4th Row: HR ground-truth.}
\label{fig:ring}
\end{figure*}

\textbf{Annealing Training} We also validated the importance of annealing training on the datasets (more intensive results are available in the supplementary materials). 
If we only trained the model with U-Nets with a discriminator (this will recover the model pix2pix \cite{isola2016image}) and
a \emph{comparator}, the training was usually very unstable and we have to do cherry-pick from results in every iteration. 
However, with the help of refining network DAEs, the performance was boosted up significantly, reflected by both the 
quantitative results summarized in Table.~\ref{tab:motion_deblur} and the qualitative results displayed in Fig.~\ref{fig:intro}. 
Additionally, we also visualized the residual images between the output of U-Nets and the final output of DAEs or the ground-truth in Fig.~\ref{fig:anneal}. The pixel level noise approximately followed a Gaussian distribution at each iteration, but with varying mean and variance,
leading to a non-stationary and time variant noise.
It was observed that the mean and variance gradually became smaller and stable at the final training stage. 
This potentially enables the DAEs to depress noise with variances of different scales added to the input,
which is different from traditional DAEs, and allows more robust training.

\subsection{Super-Resolution}

Super-resolution is another popular inverse problem in image processing, and it aims at the restoration of high frequencies to enrich details in an image based on a set of prior examples with low resolution (LR) and corresponding high resolution (HR) images. 
The degraded operator $A$ can source from quantization error, limitation of the sensor from the capturing camera, the presence of blurriness and the use of downsampling operators to reduce the image resolution for storage purposes. 
It is well known that super-resolution is ill-posed, since for each LR image the space of corresponding HR images can be very large. 

In this experiment, we first synthesized a challenge $\times 4$ downsampling operator $A$, which was a channel-wise strided convolution and may result in the \textit{ring artifact} on low resolution images (see detailed form in supplementary materials) and trained the super-resolution model on two datasets, CUB and ImageNet. Our task is trying to super-resolve images from $64\times 64$ to $256 \times 256$.  
In addition, we also compared with SRGAN \cite{ledig2016photo} in supplementary materials. 

\paragraph{Ring Artifact Remover} The difficulty of the super-resolution is highly affected by the shape and 
scale of the convolution kernel in the degradation model. Convolving an image with a wider and flatter 
convolution kernel leads to more aliasing of high frequency information in spectral domain thus leads to 
a more challenging problem. Usually, ring artifact can be resulted from this heavy aliasing on the spectral
domain. On the contrary, a narrower and sharper convolutional kernel results to less
information loss thus is less challenging. In this task, we used a flat kernel of size $7 \times 7$ 
which can cause heavy ring artifact. Fig.~\ref{fig:ring} shows the super-resolution results on held-out testing ImageNet dataset. 
In this case, the bicubic interpolation failed to remove the ring effects produced during dowmsampling.
However, our end-to-end trained \emph{InverseNet} can well tackle this problem by greatly depressing the ring effect.


\subsection{Jointly Super-Resolution and Colorization}

\begin{figure}[t]
\centering
\includegraphics[width=0.5\textwidth]{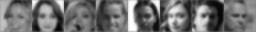}
\includegraphics[width=0.5\textwidth]{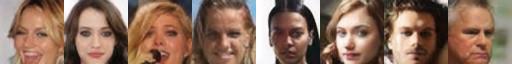}
\includegraphics[width=0.5\textwidth]{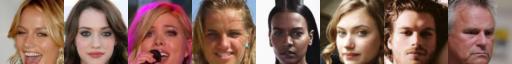}
\includegraphics[width=0.5\textwidth]{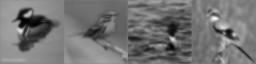}
\includegraphics[width=0.5\textwidth]{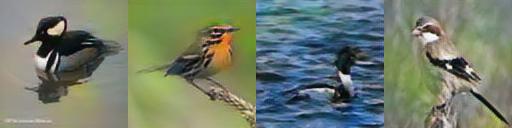}
\includegraphics[width=0.5\textwidth]{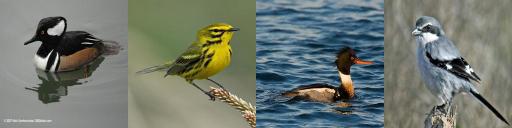}
\caption{Joint Super-resolution and colorization for CelebA and CUB datasets. (top): colorless LR images; (middle): recovery by \emph{InverseNet}; {bottom}: full color HR images.}
\label{fig:sr_color}
\end{figure}

A more challenging task is to enhance both spectral and spatial resolutions from one single band image. 
More specifically, the enhancement is a combination of super-resolution and colorization (hallucinating a plausible color version of a colorless image), which can be considered as the compound of two degraded operators. 
Thus, our single channel colorless low resolution image was obtained by convolving a kernel $A$ of size $9 \times 9 \times 3 \times 1$ and downsampling with a stride 2 in both horizontal and vertical directions. Note that this is different from and more challenging than the conventional multi-band image fusion, e.g., fusing a multispectral image and a high spatial panchromatic image (also referred to as pansharpening \cite{Loncan2015}), in the sense that only one low-spatial low-spectral resolution image is available.  
Additionally, our inverse task is also different from the traditional colorization in the CIE Lab color space. 
In this section, we have tested the ability to perform joint $\times 2$ super-resolution and colorization from one single colorless LR image on the dataset celebA and CUB. 
The celebA $32\times 32\rightarrow 64\times 64$ mainly includes faces images with less color variance, so it is an easier dataset compared with CUB wild bird images $64\times 64\rightarrow 128\times 128$. The results are displayed in Fig. \ref{fig:sr_color}.
Visually, the joint super-resolved and colorized images are very similar with their ground-truth with variances for some details.
It is especially interesting to observe that the restored high-resolution images looked more natural than the ground-truth of 
some `outlier' images. This results from the fact that the model was trained from massive images in which most look normal.
For example, the person in the third column had purple hairs which rarely appeared in the dataset. Feeding its degraded version, i.e.,
a blurred colorless image to the trained \emph{InverseNet} gave a face with most probable, popular and thus `natural' brown hairs 
as shown in the second row. More results are available in the supplementary materials.

\section{Conclusion}
In this paper we proposed the \textit{InverseNet} to solve inverse problems by end-to-end mapping. 
To take the advantage of the efficiency (in testing phase) of end-to-end learning and the flexibility
brought by splitting strategy simultaneously, the mapping was decomposed into two neural networks, 
i.e., the inversion network and the denoising one. The former one was designed to learn the inversion 
of the physical forward model associated with the data term and the latter one was to learn a proximity 
operator or a projection onto the ground-truth signal space associated with the prior term. 
The two pre-trained deep neural networks were trained jointly using prepared data pairs $(\bfx_i,\bfy_i)$,
getting rid of any two-step separate training. Experiments and analysis on various datasets
demonstrated the efficiency and accuracy of the proposed method. In future work we hope to further 
extend the proposed method to tackle the `learn to learn' problem.


\clearpage
\newpage

{\small
\bibliographystyle{ieee}
\bibliography{strings_all_ref,reference}
}

\newpage

\onecolumn

\begin{appendices}

\section{Training details}

\subsection{Adversarial training algorithm}
We define a soft GAN loss with the following form
\begin{align}
\mathcal{L}_{GAN}^{soft}(\mathbf{x}, l) = -l \cdot \log D(\mathbf{x}) - (1-l) \cdot \log (1-D(\mathbf{x}))
\end{align}
where $l\in [0,1]$ is the soft label for data $\mathbf{x}$. 
The explicit forms of the Discriminator (D) loss and the generator (G) loss are 
\begin{align}
\mathcal{L}_D &= \mathcal{L}_{GAN}^{soft}(\mathbf{x}, 0.99) + \mathcal{L}_{GAN}^{soft}(\mathbf{z}, 0.01) + \mathcal{L}_{GAN}^{soft}(\hat{\mathbf{x}}, 0.01)  \\
\mathcal{L}_G(\mathbf{z}) &= \mathcal{L}_{GAN}^{soft}(\mathbf{z}, 0.99) + \lambda_r\cdot\|\mathbf{z} - \mathbf{x} \|_2^2 + \lambda_f\cdot\|C(\mathbf{z}) - C(\mathbf{x})\|_2^2 \\
\mathcal{L}_G(\hat{\mathbf{x}}) &= \mathcal{L}_{GAN}^{soft}(\hat{\mathbf{x}}, 0.99) + \lambda_r\cdot\|\hat{\mathbf{x}} - \mathbf{x} \|_2^2 + \lambda_f\cdot\|C(\hat{\mathbf{x}}) - C(\mathbf{x})\|_2^2 
\end{align}
where $\mathbf{z}$ is the output of U-Nets, $\hat{\mathbf{x}}$ is the output of DAE and $C(\cdot)$ is the comparator output.
In our experiment, we used FC6 layer of the AlexNet. The tunable hyper-parameters $\lambda_r=\lambda_f=0.5$ were used in 
our experiment. Experimentally, this soft loss can alleviate the vulnerability of neural networks to adversarial examples.

Instead of pre-training the generator as in SRGAN, we jointly trained the whole model, i.e., the generator and discriminator simultaneously. 
The detailed optimization updates for adversarial training are summarized in Algorithm \ref{alg:InverseNet}. 
We chose $K=1$ in all experiments, and Adam \cite{kingma2014adam} optimizer was applied in the gradient descent for all parameter updates. 
For Adam, the learning rate is $10^{-4}$, and $\beta_1=0.5, \beta_2=0.999, \epsilon=10^{-8}$. 
In addition, the gradient is clipped by norm 5 for each net. 
The batch size of each iteration is 36 for all experiments. 
For batch normalization, momentum is 0.9 and $\epsilon=10^{-5}$. 

\begin{algorithm}
		\caption{Adversarial training for InverseNet}
		\label{alg:InverseNet}
		\begin{algorithmic}
		%
		\FOR{$ t = 1,2,\dots $}
		  \STATE Randomly get batch data pairs $(\mathbf{x}, \mathbf{y})$;
			\FOR{$ k = 1, 2, \dots, K$ }
				\STATE Update parameter of \textbf{Discriminator} with gradient $\nabla \mathcal{L}_D$;
     \ENDFOR
			\STATE Update parameter of \textbf{U-Nets} with gradient $\nabla \mathcal{L}_G(\mathbf{z})$;
			\STATE Update parameter of \textbf{DAEs} with gradient $\nabla \mathcal{L}_G(\hat{\mathbf{x}})$;
		\ENDFOR
		\end{algorithmic}
\end{algorithm}

\subsection{Model structure}

In Table~\ref{tab:unet}, we describe the model structure for U-Nets with input of size $256 \times 256$.
The structure of DAEs is summarized in Table~\ref{tab:dae}). 
Beside of the pixel shuffling layer, all the other layers can share the same structure as the sizes, i.e., the height and width of an input image
and an output image are the same.

\begin{table*}[h]
\centering
\caption{Network hyper-parameters of U-Nets}
\begin{tabular}{ |c|c|c| } 
\hline
Layer Name &  Dimension    & Layer  Operations                                                     \\
\hline
data  & $256 \times 256 \times 3$    & Conv($4,4,3,16$)-`SAME'                           \\
e1    & $128 \times 128 \times 16$   & Leaky Relu-Conv($4,4,16,32$)-`SAME'-Batch\_Norm   \\
e2    & $64 \times 64 \times 32$     & Leaky Relu-Conv($4,4,32,64$)-`SAME'-Batch\_Norm   \\
e3    & $32 \times 32 \times 64$     & Leaky Relu-Conv($4,4,64,128$)-`SAME'-Batch\_Norm  \\
e4    & $16 \times 16 \times 128$    & Leaky Relu-Conv($4,4,128,128$)-`SAME'-Batch\_Norm \\
e5    & $8 \times 8 \times 128$      & Leaky Relu-Conv($4,4,128,128$)-`SAME'-Batch\_Norm \\
e6    & $4 \times 4 \times 128$      & Leaky Relu-Conv($4,4,128,128$)-`SAME'-Batch\_Norm  \\
e7    & $2 \times 2 \times 128$      & Leaky Relu-Conv($4,4,128,128$)-`SAME'-Batch\_Norm  \\
e8    & $1 \times 1 \times 128$      & Relu-Conv\_Trans($4,4,128,128$)-`SAME'-Batch\_Norm-Cancat(e7)  \\\
d1    & $2 \times 2 \times 256$      & Relu-Conv\_Trans($4,4,256,128$)-`SAME'-Batch\_Norm-Cancat(e6)  \\
d2    & $4 \times 4 \times 256$      & Relu-Conv\_Trans($4,4,256,128$)-`SAME'-Batch\_Norm-Cancat(e5) \\
d3    & $8 \times 8 \times 256$      & Relu-Conv\_Trans($4,4,256,128$)-`SAME'-Batch\_Norm-Cancat(e4) \\
d4    & $16 \times 16 \times 256$    & Relu-Conv\_Trans($4,4,256,64$)-`SAME'-Batch\_Norm-Cancat(e3) \\
d5    & $32 \times 232\times 128$    & Relu-Conv\_Trans($4,4,128,32$)-`SAME'-Batch\_Norm-Cancat(e2) \\
d6    & $64 \times 64 \times 64$     & Relu-Conv\_Trans($4,4,64,16$)-`SAME'-Batch\_Norm-Cancat(e1) \\
d7    & $128 \times 1282 \times 32$  & Relu-Conv\_Trans($4,4,32,3$)-`SAME'-Tanh \\
d8    & $256 \times 256 \times 3$    & \\
\hline
\end{tabular}\label{tab:unet}
\end{table*}

\begin{table*}[h]
\centering
\caption{Network hyper-parameters of DAEs}
\begin{tabular}{ |c|c|c| } 
\hline
Input Dimension    & Layer                                           & Output Dimension \\
\hline
$256 \times 256 \times 3$      & periodical pixel shuffling                     & $64 \times  64 \times 48$ \\
$64 \times 64 \times 48$       & Conv($4,4,48,128$)-`SAME'-Batch\_Norm-Relu       & $64 \times 64 \times 128$ \\
$64 \times 64 \times 128$      & Conv($4,4,128,64$)-`SAME'-Batch\_Norm-Relu       & $64 \times 64 \times  64 $ \\
$64 \times 64 \times 64$       & Conv($4,4,64,32$)-`SAME'-Batch\_Norm-Relu        & $64 \times 64 \times 32$ \\
$64 \times 64 \times \{32,3\}$ & Concatenate in Channel                         & $64 \times 64 \times 35 $ \\
$64 \times 64 \times 35$       & Conv($4,4,35,64$)-`SAME'-Batch\_Norm-Relu        & $64 \times 64 \times 64 $\\
$64 \times 64 \times 64 $      & Conv($4,4,64,128$)-`SAME'-Batch\_Norm-Relu       & $64 \times 64 \times 128$ \\
$64 \times 64 \times 128 $     & Conv($4,4,128,48$)-`SAME'-Batch\_Norm-Relu       & $64 \times 64 \times 48$ \\
$64 \times 64 \times 48  $     & periodical pixel shuffling                     & $256 \times 256 \times 3$ \\
\hline
\end{tabular}\label{tab:dae}
\end{table*}

\section{More results for motion deblurring}

\begin{figure*}[ht]
\centering
\subfigure[Blurred Image]{
\includegraphics[width=0.23\textwidth]{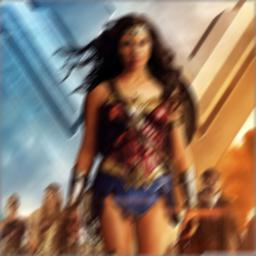}
}
\subfigure[Neural Deblur]{
\includegraphics[width=0.23\textwidth]{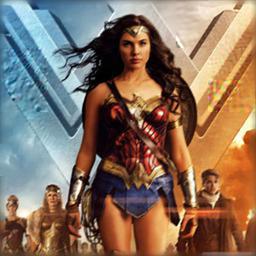}
}
\subfigure[InverseNet]{
\includegraphics[width=0.23\textwidth]{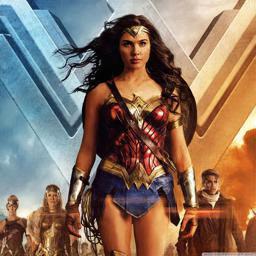}
}
\subfigure[Original Image]{
\includegraphics[width=0.23\textwidth]{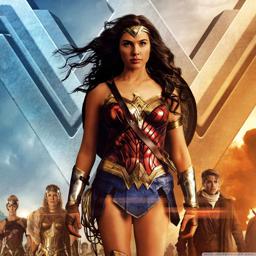}
}
\caption{Motion deblurring for a $256\times 256$ image with model trained on $128\times 128$ images.}
\label{fig:mot_deblur}
\end{figure*}

\begin{figure*}[ht]
\centering
\includegraphics[width=0.33\textwidth]{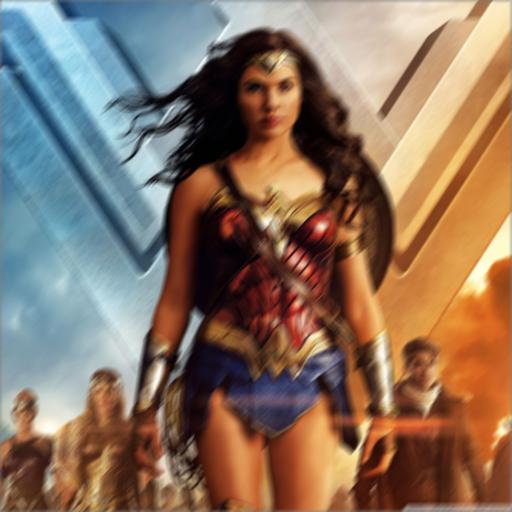}
\includegraphics[width=0.33\textwidth]{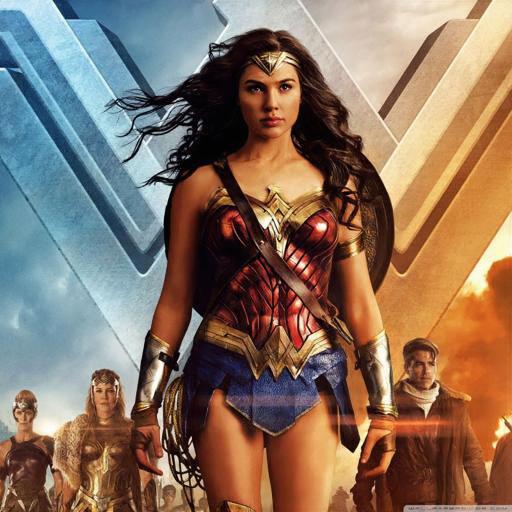}
\includegraphics[width=0.33\textwidth]{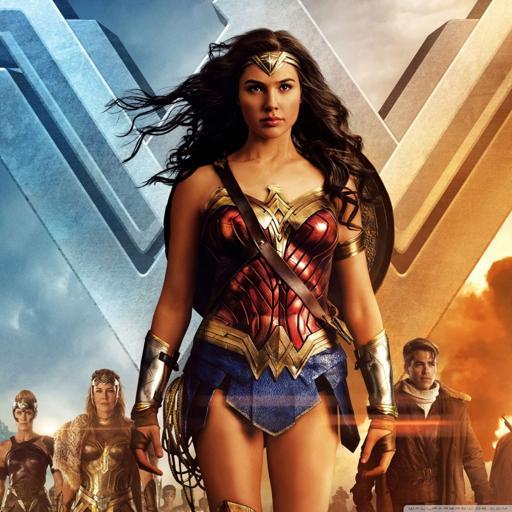}
\includegraphics[width=0.33\textwidth]{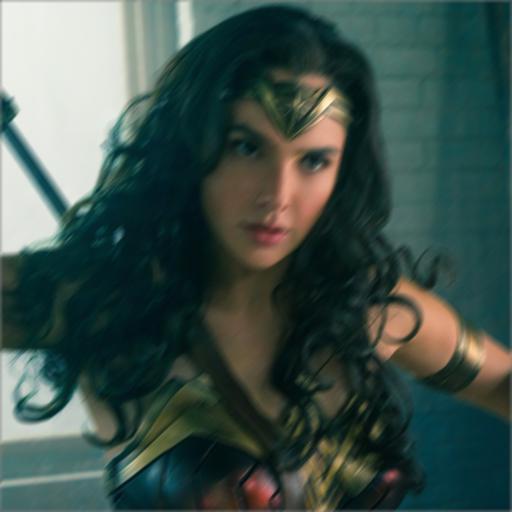}
\includegraphics[width=0.33\textwidth]{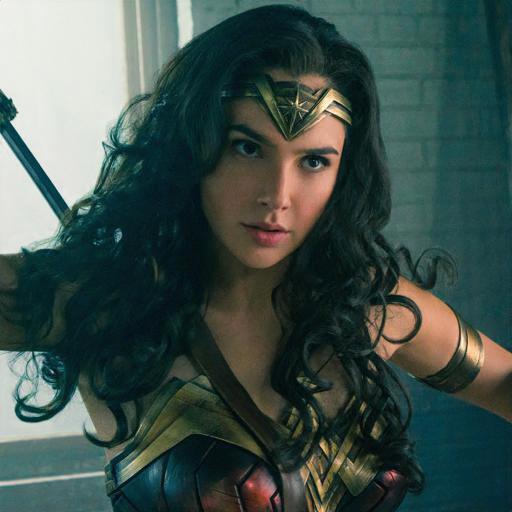}
\includegraphics[width=0.33\textwidth]{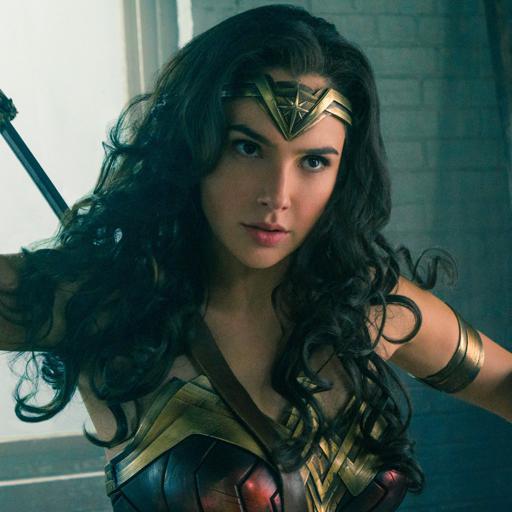}
\caption{Motion deblurring for $512\times 512$ image with model trained on $128\times 128$ images. From left to right: blurred image, deblurred image by InverseNet, and original image.}
\label{fig:mot_deblur_512}
\end{figure*}

We tested our motion deblurred model on another natural image as shown in Fig. \ref{fig:mot_deblur}. 
The model was trained on $128 \times 128$ patches from ImageNet, and tested on one Wonder Woman poster image 
with size $256\times 256$ and $512 \times 512$. 
We also compared it with the blind neural motion deblurring algorithm \cite{chakrabarti2016neural} on $256\times 256$
\footnote{We used the codes offered by the authors in \url{https://github.com/ayanc/ndeblur}.}.
As shown in Fig. \ref{fig:mot_deblur}, the deblurred image using InverseNet is much more clear and 
closer to the original image than the neural deblurred one. 
In the $512 \times 512$ size deblurring, visually, the restored image by InverseNet is almost exactly the same
with the original one as displayed in Fig. \ref{fig:mot_deblur_512}.

\begin{figure*}[h]
\centering
\subfigure[Low Resolution Images]{
\includegraphics[width=\textwidth]{figures/imagenet_lr_ring.jpg}
}
\subfigure[Bicubic]{
\includegraphics[width=\textwidth]{figures/imagenet_bicubic_ring.jpg}
}
\subfigure[SRGAN]{
\includegraphics[width=\textwidth]{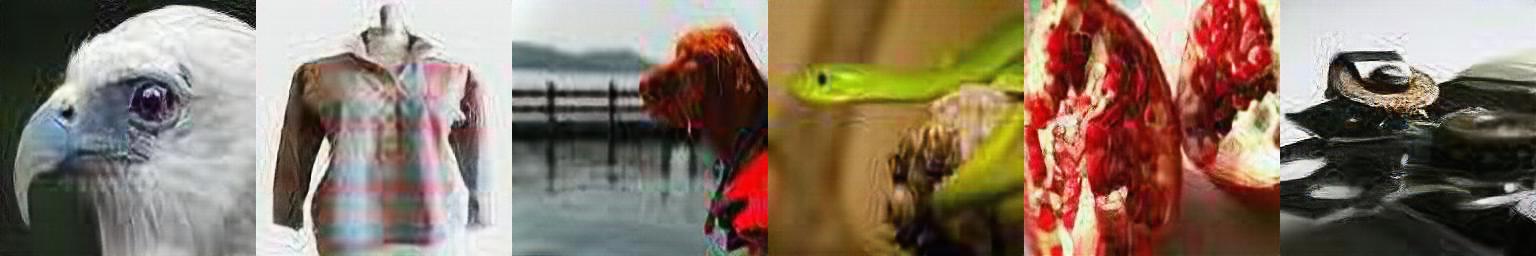}
}
\subfigure[InverseNet]{
\includegraphics[width=\textwidth]{figures/imagenet_sr_ring.jpg}
}
\subfigure[High Resolution Image]{
\includegraphics[width=\textwidth]{figures/imagenet_hr_ring.jpg}
}
\caption{\textbf{Ring Effect Remover} on ImageNet for $64\times 64\rightarrow 256\times 256$.}
\label{fig:ring_remove}
\end{figure*}

\section{More results for ring artifact remover super-resolution}
We include the results of SRGAN \cite{ledig2016photo} for comparison as shown in Fig. \ref{fig:ring_remove}. 
For SRGAN, except downgraded kernel, we follow the training instruction in the paper by first pretraining the generator
for $10^6$ iterations and then fine-tuning the model with discriminator included for another $10^6$ iterations. 
Since the author did not release the codes, this implementation was based on our own and we tried our best to fine tune these parameters
to the best performance. As shown in Fig. \ref{fig:ring_remove}, the super-resolved result by SRGAN did 
not remove the ring effect but could sharpen some details of images. On the contrary, the results of InverseNet
successfully removed the ring artifact while keeping the details.


\section{Visualization of degraded kernel $A$}

\subsection{Motion deblurring}
The degradation matrix $A$ in motion deblurring is a square matrix corresponding to a 2-D convolution.
If the convolution is implemented with periodic boundary conditions, i.e., the pixels out of an image is
padded with periodic extension of itself, the matrix $H$ is a block circulant matrix with circulant blocks (BCCB).
The first row of the matrix $A$ is the motion blurring convolution kernel and the other row vectors
are rotated one element to the right relative to the preceding row vector.
The $9 \times 9$ motion blurring convolution kernel is displayed as below.  
\begin{figure}[H]
\centering
\includegraphics[width=0.25\textwidth]{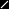}
\caption{{$9 \times 9$ motion blurring kernel.}}
\label{fig:mot_kernel}
\end{figure}

\subsection{Super-resolution}
\label{subsec:sr}
The degradation matrix $A$ in super-resolution can be decomposed as the multiplication of  a 
convolution matrix $H$ (as in motion blurring) and a down-sampling matrix $S$, i.e., $A= SH$. 
The down-sampling matrix $S$ represents the regular 2-D decimation by keeping one pixel every
$d$ pixels in both horizontal and vertical directions, where $d$ is the sampling ratio. 
The matrix $A$ is equivalent to the strided convolution widely used in CNN architecture. 
The $3 \times 3$ convolution kernel used in our experiment is displayed as below.
The down-sampling ratio $d$ is fixed to 4 in the experiments, corresponding to shrinking the size
of an image from $256 \times 256$ to $64 \times 64$.

\begin{figure}[H]
\centering
\includegraphics[width=0.3\textwidth]{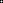}
\caption{{$3 \times 3$ convolution kernel in super-resolution.}}
\label{fig:sr_kernel}
\end{figure}

\subsection{Joint super-resolution and colorization}
\begin{figure}[H]
\centering
\includegraphics[width=0.3\textwidth]{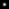}
\includegraphics[width=0.3\textwidth]{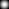}
\includegraphics[width=0.3\textwidth]{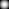}
\caption{{Three $9 \times 9$ convolution kernels for red (left), green (middle) and blue (right) channels.}}
\label{fig:joint_kernel}
\end{figure}

The degradation matrix $A$ in joint super-resolution and colorization can be decomposed as 
the product of a convolution matrix $H$, a down-sampling matrix $S$ and a spectral degradation matrix $L$.
The matrices $H$ and $S$ are similarly defined as in Section \ref{subsec:sr}. The main difference here
is that the kernel of $H$ is channel wise in the sense that the convolution kernel for each channel is 
different, as shown in Fig. \ref{fig:joint_kernel}. The role of spectral degradation $L$ is making the average
of the R, G, B channels to get a one-channel grayscale image.

\end{appendices}

\end{document}

%% file: notations.tex
\input com_notations.tex

\newcommand{\apriori}{\emph{a priori }}
\newcommand{\aposteriori}{\emph{a posteriori }}

\newcommand{\ALLobs}{\boldsymbol{\calY}}
\newcommand{\allobs}{\boldsymbol{\mathtt{z}}}
\newcommand{\MATobs}[1]{\bfZ_{#1}}
\newcommand{\Vobs}[1]{\bfz_{#1}}

\newcommand{\MATima}{\bfX}
\newcommand{\Vima}{\bfx}
\newcommand{\ima}[1]{x_{#1}}

\newcommand{\ftrans}[2]{\calF_{#1}\left(#2\right)}
\newcommand{\MATtrans}[1]{\bfF_{#1}}
\newcommand{\kernel}[1]{\boldsymbol{\kappa}_{#1}}

\newcommand{\MATnoise}[1]{\bfE_{#1}}
\newcommand{\Vnoise}[1]{\bfe_{#1}}
\newcommand{\noisevar}[1]{{s^2_{#1}}}
\newcommand{\Vnoisevar}{\bss^2}

\newcommand{\noobs}{p}
\newcommand{\nbobs}{P}

\newcommand{\nbrowobs}[1]{n_{\mathrm{x},#1}}
\newcommand{\nbcolobs}[1]{n_{\mathrm{y},#1}}
\newcommand{\nbbandobs}[1]{n_{\lambda,#1}}
\newcommand{\nbpixobs}[1]{N_{#1}}

\newcommand{\nbrowima}{m_{\mathrm{x}}}
\newcommand{\nbcolima}{m_{\mathrm{y}}}
\newcommand{\nbbandima}{m_{\lambda}}
\newcommand{\nbpixima}{M}

\newcommand{\imam}[1]{\boldsymbol{\mu}_{#1}}
\newcommand{\meansub}{\imam{\bsu}}
\newcommand{\Covsub}{\boldsymbol{\Sigma}}
\newcommand{\imamall}{\imam{\bfu}^{\star}}
\newcommand{\imacovall}{\imacovmat{\bfu}^{\star}}

\newcommand{\imacovmat}[1]{\boldsymbol{\Sigma}_{#1}}
\newcommand{\covmat}[1]{\boldsymbol{\Sigma}_{#1}} 
\newcommand{\CovEstHS}{\hat{\boldsymbol{\Sigma}}_{\bsz_{1}}}

\newcommand{\imacoefgm}{\alpha}
\newcommand{\imahid}{\boldsymbol{\zeta}}
\newcommand{\nei}{\boldsymbol{\nu}}

\newcommand{\hypervect}{\boldsymbol{\Phi}}
\newcommand{\paramvect}{\boldsymbol{\theta}}
%
%
\newcommand{\sample}[2]{\tilde{#1}^{#2}}
\newcommand{\samplebis}[2]{{#1}^{\left(#2\right)}}
\newcommand{\samplenoisevar}[1]{{\widetilde{\sigma}}^{2(#1)}}
\newcommand{\sampleparamvect}[1]{\widetilde{\paramvect}^{(#1)}}

\newcommand{\MAP}[1]{\hat{#1}_{\mathrm{MAP}}}
\newcommand{\MMSE}[1]{\hat{#1}_{\mathrm{MMSE}}}
\newcommand{\argmax}{\mathrm{arg}\max}
\newcommand{\argmin}{\mathrm{arg}\min}

\newcommand{\norm}[1]{\left\|#1\right\|}

\newcommand{\R}{\mathds{R}}

\newcommand{\dirac}[1]{\delta\left({#1}\right)}

\newcommand{\inv}{^{-1}}

\newcommand{\herm}{^{*}}

\newcommand{\transp}{^T}

\newcommand{\etr}{\mathrm{etr}}

\newcommand{\Ndistr}[1]{\mathcal{N}\left(#1\right)}

\newcommand{\Vun}{{\boldsymbol{1}}}
\newcommand{\Vzero}[1]{\boldsymbol{0}_{#1}}
\newcommand{\Vzeros}[1]{\boldsymbol{0}_{#1}}
\newcommand{\Id}[1]{\textbf{I}_{#1}}
\newcommand{\Indicfun}[2]{\textbf{1}_{#1}\left(#2\right)}
\newcommand{\Diag}[2]{\left[#1\right]_{#2}}

\newenvironment{algogo}[1]{
\smallskip
\noindent \hrule\vspace{0.2\baselineskip} \hrule
\smallskip
\begin{small}
\refstepcounter{algo} \center{\bf \textsc{Algorithm \thealgo:}}
\\{\center{\bf #1}}
\smallskip
\flushleft
 } {
\end{small}
\bigskip
\hrule\vspace{0.2\baselineskip} \hrule
\smallskip }

\newcounter{algo}
\renewcommand{\thealgo}{\arabic{algo}}

%% file: com_notations.tex

\def\bfa{{\mathbf{a}}}
\def\bfb{{\mathbf{b}}}
\def\bfc{{\mathbf{c}}}
\def\bfd{{\mathbf{d}}}
\def\bfe{{\mathbf{e}}}
\def\bff{{\mathbf{f}}}
\def\bfg{{\mathbf{g}}}
\def\bfh{{\mathbf{h}}}
\def\bfi{{\mathbf{i}}}
\def\bfj{{\mathbf{j}}}
\def\bfk{{\mathbf{k}}}
\def\bfl{{\mathbf{l}}}
\def\bfm{{\mathbf{m}}}
\def\bfn{{\mathbf{n}}}
\def\bfo{{\mathbf{o}}}
\def\bfp{{\mathbf{p}}}
\def\bfq{{\mathbf{q}}}
\def\bfr{{\mathbf{r}}}
\def\bfs{{\mathbf{s}}}
\def\bft{{\mathbf{t}}}
\def\bfu{{\mathbf{u}}}
\def\bfv{{\mathbf{v}}}
\def\bfw{{\mathbf{w}}}
\def\bfx{{\mathbf{x}}}
\def\bfy{{\mathbf{y}}}
\def\bfz{{\mathbf{z}}}

\def\bfA{{\mathbf{A}}}
\def\bfB{{\mathbf{B}}}
\def\bfC{{\mathbf{C}}}
\def\bfD{{\mathbf{D}}}
\def\bfE{{\mathbf{E}}}
\def\bfF{{\mathbf{F}}}
\def\bfG{{\mathbf{G}}}
\def\bfH{{\mathbf{H}}}
\def\bfI{{\mathbf{I}}}
\def\bfJ{{\mathbf{J}}}
\def\bfK{{\mathbf{K}}}
\def\bfL{{\mathbf{L}}}
\def\bfM{{\mathbf{M}}}
\def\bfN{{\mathbf{N}}}
\def\bfO{{\mathbf{O}}}
\def\bfP{{\mathbf{P}}}
\def\bfQ{{\mathbf{Q}}}
\def\bfR{{\mathbf{R}}}
\def\bfS{{\mathbf{S}}}
\def\bfT{{\mathbf{T}}}
\def\bfU{{\mathbf{U}}}
\def\bfV{{\mathbf{V}}}
\def\bfW{{\mathbf{W}}}
\def\bfX{{\mathbf{X}}}
\def\bfY{{\mathbf{Y}}}
\def\bfZ{{\mathbf{Z}}}


\def\bbA{{\mathbb{A}}}
\def\bbB{{\mathbb{B}}}
\def\bbC{{\mathbb{C}}}
\def\bbD{{\mathbb{D}}}
\def\bbE{{\mathbb{E}}}
\def\bbF{{\mathbb{F}}}
\def\bbG{{\mathbb{G}}}
\def\bbH{{\mathbb{H}}}
\def\bbI{{\mathbb{I}}}
\def\bbJ{{\mathbb{J}}}
\def\bbK{{\mathbb{K}}}
\def\bbL{{\mathbb{L}}}
\def\bbM{{\mathbb{M}}}
\def\bbN{{\mathbb{N}}}
\def\bbO{{\mathbb{O}}}
\def\bbP{{\mathbb{P}}}
\def\bbQ{{\mathbb{Q}}}
\def\bbR{{\mathbb{R}}}
\def\bbS{{\mathbb{S}}}
\def\bbT{{\mathbb{T}}}
\def\bbU{{\mathbb{U}}}
\def\bbV{{\mathbb{V}}}
\def\bbW{{\mathbb{W}}}
\def\bbX{{\mathbb{X}}}
\def\bbY{{\mathbb{Y}}}
\def\bbZ{{\mathbb{Z}}}


\def\dsA{{\mathds{A}}}
\def\dsB{{\mathds{B}}}
\def\dsC{{\mathds{C}}}
\def\dsD{{\mathds{D}}}
\def\dsE{{\mathds{E}}}
\def\dsF{{\mathds{F}}}
\def\dsG{{\mathds{G}}}
\def\dsH{{\mathds{H}}}
\def\dsI{{\mathds{I}}}
\def\dsJ{{\mathds{J}}}
\def\dsK{{\mathds{K}}}
\def\dsL{{\mathds{L}}}
\def\dsM{{\mathds{M}}}
\def\dsN{{\mathds{N}}}
\def\dsO{{\mathds{O}}}
\def\dsP{{\mathds{P}}}
\def\dsQ{{\mathds{Q}}}
\def\dsR{{\mathds{R}}}
\def\dsS{{\mathds{S}}}
\def\dsT{{\mathds{T}}}
\def\dsU{{\mathds{U}}}
\def\dsV{{\mathds{V}}}
\def\dsW{{\mathds{W}}}
\def\dsX{{\mathds{X}}}
\def\dsY{{\mathds{Y}}}
\def\dsZ{{\mathds{Z}}}
\def\dsz{{\mathds{z}}}

\def\calh{{\mathcal{h}}}
\def\calU{{\mathcal{U}}}
\def\calu{{\mathcal{u}}}
\def\calS{{\mathcal{S}}}
\def\caln{{\mathcal{n}}}
\def\calV{{\mathcal{V}}}
\def\calv{{\mathcal{v}}}
\def\calP{{\mathcal{P}}}
\def\calA{{\mathcal{A}}}
\def\calB{{\mathcal{B}}}
\def\calC{{\mathcal{C}}}
\def\calD{{\mathcal{D}}}
\def\calE{{\mathcal{E}}}
\def\calF{{\mathcal{F}}}
\def\calG{{\mathcal{G}}}
\def\calH{{\mathcal{H}}}
\def\calI{{\mathcal{I}}}
\def\calJ{{\mathcal{J}}}
\def\calK{{\mathcal{K}}}
\def\calL{{\mathcal{L}}}
\def\calM{{\mathcal{M}}}
\def\calN{{\mathcal{N}}}
\def\calO{{\mathcal{O}}}
\def\calP{{\mathcal{P}}}
\def\calQ{{\mathcal{Q}}}
\def\calR{{\mathcal{R}}}
\def\calS{{\mathcal{S}}}
\def\calT{{\mathcal{T}}}
\def\calU{{\mathcal{U}}}
\def\calV{{\mathcal{V}}}
\def\calW{{\mathcal{W}}}
\def\calx{{\mathcal{x}}}
\def\calX{{\mathcal{X}}}
\def\caly{{\mathcal{y}}}
\def\calY{{\mathcal{Y}}}
\def\calZ{{\mathcal{Z}}}
\def\calz{{\mathcal{z}}}


\def\bsa{{\boldsymbol{a}}}
\def\bsb{{\boldsymbol{b}}}
\def\bsc{{\boldsymbol{c}}}
\def\bsd{{\boldsymbol{d}}}
\def\bse{{\boldsymbol{e}}}
\def\bsf{{\boldsymbol{f}}}
\def\bsg{{\boldsymbol{g}}}
\def\bsh{{\boldsymbol{h}}}
\def\bsi{{\boldsymbol{i}}}
\def\bsj{{\boldsymbol{j}}}
\def\bsk{{\boldsymbol{k}}}
\def\bsl{{\boldsymbol{l}}}
\def\bsm{{\boldsymbol{m}}}
\def\bsn{{\boldsymbol{n}}}
\def\bso{{\boldsymbol{o}}}
\def\bsp{{\boldsymbol{p}}}
\def\bsq{{\boldsymbol{q}}}
\def\bsr{{\boldsymbol{r}}}
\def\bss{{\boldsymbol{s}}}
\def\bst{{\boldsymbol{t}}}
\def\bsu{{\boldsymbol{u}}}
\def\bsv{{\boldsymbol{v}}}
\def\bsw{{\boldsymbol{w}}}
\def\bsx{{\boldsymbol{x}}}
\def\bsy{{\boldsymbol{y}}}
\def\bsz{{\boldsymbol{z}}}

\def\bsA{{\boldsymbol{A}}}
\def\bsB{{\boldsymbol{B}}}
\def\bsC{{\boldsymbol{C}}}
\def\bsD{{\boldsymbol{D}}}
\def\bsE{{\boldsymbol{E}}}
\def\bsF{{\boldsymbol{F}}}
\def\bsG{{\boldsymbol{G}}}
\def\bsH{{\boldsymbol{H}}}
\def\bsI{{\boldsymbol{I}}}
\def\bsJ{{\boldsymbol{J}}}
\def\bsK{{\boldsymbol{K}}}
\def\bsL{{\boldsymbol{L}}}
\def\bsM{{\boldsymbol{M}}}
\def\bsN{{\boldsymbol{N}}}
\def\bsO{{\boldsymbol{O}}}
\def\bsP{{\boldsymbol{P}}}
\def\bsQ{{\boldsymbol{Q}}}
\def\bsR{{\boldsymbol{R}}}
\def\bsS{{\boldsymbol{S}}}
\def\bsT{{\boldsymbol{T}}}
\def\bsU{{\boldsymbol{U}}}
\def\bsV{{\boldsymbol{V}}}
\def\bsW{{\boldsymbol{W}}}
\def\bsX{{\boldsymbol{X}}}
\def\bsY{{\boldsymbol{Y}}}
\def\bsZ{{\boldsymbol{Z}}}

\def\wtm{\widetilde{m}}
\def\wtM{\widetilde{M}}
\def\wtV{\widetilde{\bfV}}
\def\wtR{\widetilde{\bfR}}
\def\whs{\widehat{s}}
\def\mfV{\mathfrak{V}}

%% file: Neural_ADMM.bbl
\begin{thebibliography}{10}\itemsep=-1pt

\bibitem{adler2017solving}
J.~Adler and O.~{\"O}ktem.
\newblock Solving ill-posed inverse problems using iterative deep neural
  networks.
\newblock {\em arXiv preprint arXiv:1704.04058}, 2017.

\bibitem{Afonso2010}
M.~V. Afonso, J.~M. Bioucas-Dias, and M.~A. Figueiredo.
\newblock Fast image recovery using variable splitting and constrained
  optimization.
\newblock {\em IEEE Trans. Image Process.}, 19(9):2345--2356, 2010.

\bibitem{arjovsky2017wasserstein}
M.~Arjovsky, S.~Chintala, and L.~Bottou.
\newblock Wasserstein gan.
\newblock {\em arXiv preprint arXiv:1701.07875}, 2017.

\bibitem{Boyd2011}
S.~Boyd, N.~Parikh, E.~Chu, B.~Peleato, and J.~Eckstein.
\newblock Distributed optimization and statistical learning via the alternating
  direction method of multipliers.
\newblock {\em Foundations and Trends{\textregistered} in Machine Learning},
  3(1):1--122, 2011.

\bibitem{bruna2015super}
J.~Bruna, P.~Sprechmann, and Y.~LeCun.
\newblock Super-resolution with deep convolutional sufficient statistics.
\newblock {\em arXiv preprint arXiv:1511.05666}, 2015.

\bibitem{chakrabarti2016neural}
A.~Chakrabarti.
\newblock A neural approach to blind motion deblurring.
\newblock In {\em European Conference on Computer Vision}, pages 221--235.
  Springer, 2016.

\bibitem{chang2017one}
J.~Chang, C.-L. Li, B.~Poczos, B.~Kumar, and A.~C. Sankaranarayanan.
\newblock One network to solve them all---solving linear inverse problems using
  deep projection models.
\newblock {\em arXiv preprint arXiv:1703.09912}, 2017.

\bibitem{chen2017photographic}
Q.~Chen and V.~Koltun.
\newblock Photographic image synthesis with cascaded refinement networks.
\newblock {\em arXiv preprint arXiv:1707.09405}, 2017.

\bibitem{cciccek20163d}
{\"O}.~{\c{C}}i{\c{c}}ek, A.~Abdulkadir, S.~S. Lienkamp, T.~Brox, and
  O.~Ronneberger.
\newblock 3d u-net: learning dense volumetric segmentation from sparse
  annotation.
\newblock In {\em International Conference on Medical Image Computing and
  Computer-Assisted Intervention}, pages 424--432. Springer, 2016.

\bibitem{Combettes2011}
P.~Combettes and J.-C. Pesquet.
\newblock Proximal splitting methods in signal processing.
\newblock In H.~H. Bauschke, R.~S. Burachik, P.~L. Combettes, V.~Elser, D.~R.
  Luke, and H.~Wolkowicz, editors, {\em Fixed-Point Algorithms for Inverse
  Problems in Science and Engineering}, Springer Optimization and Its
  Applications, pages 185--212. Springer New York, 2011.

\bibitem{Csaji2001}
B.~C. Cs{\'a}ji.
\newblock Approximation with artificial neural networks.
\newblock {\em Faculty of Sciences, Etvs Lornd University, Hungary}, 24:48,
  2001.

\bibitem{deng2009imagenet}
J.~Deng, W.~Dong, R.~Socher, L.-J. Li, K.~Li, and L.~Fei-Fei.
\newblock Imagenet: A large-scale hierarchical image database.
\newblock In {\em Computer Vision and Pattern Recognition, 2009. CVPR 2009.
  IEEE Conference on}, pages 248--255. IEEE, 2009.

\bibitem{Deng2014deep}
L.~Deng, D.~Yu, et~al.
\newblock Deep learning: methods and applications.
\newblock {\em Foundations and Trends{\textregistered} in Signal Processing},
  7(3--4):197--387, 2014.

\bibitem{Dong2016PAMI}
C.~Dong, C.~C. Loy, K.~He, and X.~Tang.
\newblock Image super-resolution using deep convolutional networks.
\newblock {\em IEEE Transactions on Pattern Analysis and Machine Intelligence},
  38(2):295--307, Feb 2016.

\bibitem{Elad2006}
M.~Elad and M.~Aharon.
\newblock Image denoising via sparse and redundant representations over learned
  dictionaries.
\newblock {\em IEEE Trans. Image Process.}, 15(12):3736--3745, 2006.

\bibitem{everingham2010pascal}
M.~Everingham, L.~Van~Gool, C.~K. Williams, J.~Winn, and A.~Zisserman.
\newblock The pascal visual object classes (voc) challenge.
\newblock {\em International journal of computer vision}, 88(2):303--338, 2010.

\bibitem{ADMM2017NIPS}
K.~Fan$^{\ast}$, Q.~Wei$^{\ast}$, L.~Carin, and K.~Heller.
\newblock An inner-loop free solution to inverse problems using deep neural
  networks.
\newblock In {\em Advances in Neural Information Processing Systems}, Long
  Beach, CA, USA, Dec. 2017.

\bibitem{Figueiredo2007gradient}
M.~A. Figueiredo, R.~D. Nowak, and S.~J. Wright.
\newblock Gradient projection for sparse reconstruction: Application to
  compressed sensing and other inverse problems.
\newblock {\em IEEE J. Sel. Topics Signal Process.}, 1(4):586--597, 2007.

\bibitem{gatys2015neural}
L.~A. Gatys, A.~S. Ecker, and M.~Bethge.
\newblock A neural algorithm of artistic style.
\newblock {\em arXiv preprint arXiv:1508.06576}, 2015.

\bibitem{goodfellow2014generative}
I.~Goodfellow, J.~Pouget-Abadie, M.~Mirza, B.~Xu, D.~Warde-Farley, S.~Ozair,
  A.~Courville, and Y.~Bengio.
\newblock Generative adversarial nets.
\newblock In {\em Advances in Neural Information Processing Systems}, pages
  2672--2680, 2014.

\bibitem{gregor2010learning}
K.~Gregor and Y.~LeCun.
\newblock Learning fast approximations of sparse coding.
\newblock In {\em Proceedings of the 27th International Conference on Machine
  Learning (ICML-10)}, pages 399--406, 2010.

\bibitem{He_2016_CVPR}
K.~He, X.~Zhang, S.~Ren, and J.~Sun.
\newblock Deep residual learning for image recognition.
\newblock In {\em Proc. IEEE Int. Conf. Comp. Vision and Pattern Recognition
  (CVPR)}, June 2016.

\bibitem{isola2016image}
P.~Isola, J.-Y. Zhu, T.~Zhou, and A.~A. Efros.
\newblock Image-to-image translation with conditional adversarial networks.
\newblock {\em arXiv preprint arXiv:1611.07004}, 2016.

\bibitem{Jin2017}
K.~H. Jin, M.~T. McCann, E.~Froustey, and M.~Unser.
\newblock Deep convolutional neural network for inverse problems in imaging.
\newblock {\em IEEE Trans. Image Process.}, 26(9):4509--4522, Sept 2017.

\bibitem{krizhevsky2012imagenet}
A.~Krizhevsky, I.~Sutskever, and G.~E. Hinton.
\newblock Imagenet classification with deep convolutional neural networks.
\newblock In {\em Advances in Neural Information Processing Systems}, pages
  1097--1105, 2012.

\bibitem{larsen2016autoencoding}
A.~B.~L. Larsen, S.~K. S{\o}nderby, H.~Larochelle, and O.~Winther.
\newblock Autoencoding beyond pixels using a learned similarity metric.
\newblock In {\em Proceedings of the 33rd International Conference on
  International Conference on Machine Learning-Volume 48}, pages 1558--1566.
  JMLR. org, 2016.

\bibitem{larsson2016learning}
G.~Larsson, M.~Maire, and G.~Shakhnarovich.
\newblock Learning representations for automatic colorization.
\newblock In {\em Proc. European Conf. Comp. Vision (ECCV)}, pages 577--593.
  Springer, 2016.

\bibitem{ledig2016photo}
C.~Ledig, L.~Theis, F.~Husz{\'a}r, J.~Caballero, A.~Cunningham, A.~Acosta,
  A.~Aitken, A.~Tejani, J.~Totz, Z.~Wang, et~al.
\newblock Photo-realistic single image super-resolution using a generative
  adversarial network.
\newblock {\em arXiv preprint arXiv:1609.04802}, 2016.

\bibitem{liu2015deep}
Z.~Liu, P.~Luo, X.~Wang, and X.~Tang.
\newblock Deep learning face attributes in the wild.
\newblock In {\em Proc. IEEE Int. Conf. Comp. Vision (ICCV)}, pages 3730--3738,
  2015.

\bibitem{Loncan2015}
L.~Loncan, L.~B. Almeida, {J. M. Bioucas-Dias}, X.~Briottet, J.~Chanussot,
  N.~Dobigeon, S.~Fabre, W.~Liao, G.~Licciardi, M.~Simoes, J.-Y. Tourneret,
  M.~Veganzones, G.~Vivone, Q.~Wei, and N.~Yokoya.
\newblock Hyperspectral pansharpening: a review.
\newblock {\em IEEE Geosci. Remote Sens. Mag.}, 3(3):27--46, Sept. 2015.

\bibitem{7879849}
S.~Lu, M.~Hong, and Z.~Wang.
\newblock A nonconvex splitting method for symmetric nonnegative matrix
  factorization: Convergence analysis and optimality.
\newblock {\em IEEE Transactions on Signal Processing}, 65(12):3120--3135, June
  2017.

\bibitem{Yao2004SR}
Y.~Lu, M.~Inamura, and M.~del Carmen~Valdes.
\newblock Super-resolution of the undersampled and subpixel shifted image
  sequence by a neural network.
\newblock {\em International Journal of Imaging Systems and Technology},
  14(1):8--15, 2004.

\bibitem{ronneberger2015u}
O.~Ronneberger, P.~Fischer, and T.~Brox.
\newblock U-net: Convolutional networks for biomedical image segmentation.
\newblock In {\em International Conference on Medical Image Computing and
  Computer-Assisted Intervention}, pages 234--241. Springer, 2015.

\bibitem{salimans2016improved}
T.~Salimans, I.~Goodfellow, W.~Zaremba, V.~Cheung, A.~Radford, and X.~Chen.
\newblock Improved techniques for training gans.
\newblock In {\em Advances in Neural Information Processing Systems}, pages
  2234--2242, 2016.

\bibitem{schlemper2017deep}
J.~Schlemper, J.~Caballero, J.~V. Hajnal, A.~Price, and D.~Rueckert.
\newblock A deep cascade of convolutional neural networks for {MR} image
  reconstruction.
\newblock {\em arXiv preprint arXiv:1703.00555}, 2017.

\bibitem{shankar2016refining}
S.~Shankar, D.~Robertson, Y.~Ioannou, A.~Criminisi, and R.~Cipolla.
\newblock Refining architectures of deep convolutional neural networks.
\newblock In {\em Proc. IEEE Int. Conf. Comp. Vision and Pattern Recognition
  (CVPR)}, pages 2212--2220, Las Vegas, NV, USA, 2016.

\bibitem{shi2016real}
W.~Shi, J.~Caballero, F.~Husz{\'a}r, J.~Totz, A.~P. Aitken, R.~Bishop,
  D.~Rueckert, and Z.~Wang.
\newblock Real-time single image and video super-resolution using an efficient
  sub-pixel convolutional neural network.
\newblock In {\em Proc. IEEE Int. Conf. Comp. Vision and Pattern Recognition
  (CVPR)}, pages 1874--1883, 2016.

\bibitem{Simoes2015}
M.~Simoes, J.~Bioucas-Dias, L.~Almeida, and J.~Chanussot.
\newblock A convex formulation for hyperspectral image superresolution via
  subspace-based regularization.
\newblock {\em IEEE Trans. Geosci. Remote Sens.}, 53(6):3373--3388, Jun. 2015.

\bibitem{simonyan2014very}
K.~Simonyan and A.~Zisserman.
\newblock Very deep convolutional networks for large-scale image recognition.
\newblock {\em arXiv preprint arXiv:1409.1556}, 2014.

\bibitem{sonderby2016amortised}
C.~K. S{\o}nderby, J.~Caballero, L.~Theis, W.~Shi, and F.~Husz{\'a}r.
\newblock Amortised {MAP} inference for image super-resolution.
\newblock {\em arXiv preprint arXiv:1610.04490}, 2016.

\bibitem{Tarantola2005inverse}
A.~Tarantola.
\newblock {\em Inverse problem theory and methods for model parameter
  estimation}.
\newblock SIAM, 2005.

\bibitem{Tikhonov1977}
A.~Tikhonov and V.~Arsenin.
\newblock {\em Solutions of ill-posed problems}.
\newblock Scripta series in mathematics. Winston, 1977.

\bibitem{Vincent2008}
P.~Vincent, H.~Larochelle, Y.~Bengio, and P.-A. Manzagol.
\newblock Extracting and composing robust features with denoising autoencoders.
\newblock In {\em Proc. Int. Conf. Machine Learning (ICML)}, pages 1096--1103.
  ACM, 2008.

\bibitem{wah2011caltech}
C.~Wah, S.~Branson, P.~Welinder, P.~Perona, and S.~Belongie.
\newblock The caltech-ucsd birds-200-2011 dataset.
\newblock 2011.

\bibitem{Wei2015JSTSP}
Q.~Wei, N.~Dobigeon, and J.-Y. Tourneret.
\newblock Bayesian fusion of multi-band images.
\newblock {\em IEEE J. Sel. Topics Signal Process.}, 9(6):1117--1127, Sept.
  2015.

\bibitem{xu2010two}
L.~Xu and J.~Jia.
\newblock Two-phase kernel estimation for robust motion deblurring.
\newblock In {\em European Conference on Computer Vision}, pages 157--170.
  Springer, 2010.

\bibitem{Yang2010}
J.~Yang, J.~Wright, T.~S. Huang, and Y.~Ma.
\newblock Image super-resolution via sparse representation.
\newblock {\em IEEE Trans. Image Process.}, 19(11):2861--2873, 2010.

\bibitem{Zhao2016}
N.~Zhao, Q.~Wei, A.~Basarab, N.~Dobigeon, D.~Kouam\'e, and J.~Y. Tourneret.
\newblock Fast single image super-resolution using a new analytical solution
  for $\ell_2-\ell_2$ problems.
\newblock {\em IEEE Trans. Image Process.}, 25(8):3683--3697, Aug. 2016.

\end{thebibliography}
